# Clinical value claims in multimodal AI outpace supporting evidence

Jiayu Feng[1,*]

[1] Department of Health Policy and Management, Harvard T.H. Chan School of Public Health, Boston, MA, USA

[*]Correspondence: jiayufeng@hsph.harvard.edu

## Abstract

Multimodal prediction models add a data source, such as imaging or an assay, to clinical information, and health systems must decide whether that added test is worth acquiring. We audited 177 studies comparing multimodal with unimodal models: 81 from a systematic review and 96 from a probability sample of 2025–2026 PubMed records. Gains in the area under the receiver operating characteristic curve (AUROC) were smaller when a study fitted a separate model for every data source (adjusted difference, −0.030; 95% CI, −0.056 to −0.005). Within 78 such studies, leaving out the strongest source's model would have raised the average gain from 0.046 to 0.108. This complete benchmarking fell from 69.1% of review studies to an estimated 46.0% of recent ones. An estimated 22% assessed decision utility and none evaluated a model prospectively, yet most claimed potential clinical usefulness or improved care (84.0% and an estimated 87.7%).

A multimodal clinical prediction model asks more of a care pathway than a single-source model does. Someone has to order the extra scan or assay, and the result has to arrive before the decision the model is meant to inform. Studies of these models usually support the addition with a comparison of discrimination: the combined model reaches a higher AUROC than a model built on one data source[1–3]. That comparison shows that the added source carries information the comparator lacked. A clinician ordering the test, or a health system paying for it, also needs to know whether that information is worth acquiring.

Clinical prediction research studies this question under the heading of incremental value[4,5]. The gain attributed to a new predictor depends on what it is added to: the same marker produces a smaller increase in AUROC on top of a strong baseline model than on top of a weak one, so incremental value is not a fixed property of the marker[6–8]. A study that measures its gain against a weak or incomplete set of single-source models can credit the added source with more than it contributes. Discrimination is also a weak guide to clinical consequence[8]. Whether a new test changes decisions depends on calibration[9], on the decision thresholds in use and on what acting costs, and is judged with decision-analytic measures such as net benefit[10,11] and, in the end, by prospective evaluation[12]. For a data source that must be acquired, the burden of obtaining it also matters[2,13].

A recent systematic review of 97 studies found that multimodal models outperformed unimodal ones in 91% of cases[14]. We asked how far the published evidence goes beyond that finding. We re-coded the studies in that review and a stratified probability sample of studies indexed since, and selected one primary multimodal-versus-unimodal comparison per study with a hierarchy that never selected the multimodal model by its score. For each study we recorded whether every constituent data source had its own separately fitted model and whether the gain came with an uncertainty estimate and was confirmed on external data. We also recorded whether calibration, decision utility and acquisition burden were reported, whether the model had been evaluated prospectively, and how strongly the authors described its clinical value.

## Results

### *Multimodal models usually outperformed their unimodal comparators*

The analysis included 177 studies. Of the 97 legacy studies, full text was obtained for 95 and 81 met the eligibility criteria. In the update period, 170 of 926 PubMed records were sampled across three title-and-abstract screening strata (Methods). Full text was unavailable for 16, three records duplicated legacy studies and one preprint had been superseded by its peer-reviewed version. Of the remaining 150, 96 were eligible (83, 8 and 5 from the three strata). The design-based estimate of eligible studies in the update frame was 434 (s.e. 39) of 926 records. Legacy figures below are proportions of a census. Update figures are design-weighted estimates for the 926-record frame, labelled as estimates, with unweighted counts in parentheses where useful.

The corpora were similar in modality mix and specialty (Extended Data Table 1). Most primary multimodal models included an imaging study, structured clinical data or both. Imaging was one of the inputs in 88.9% of legacy studies and an estimated 85.4% of update studies, and structured clinical or electronic health record data in 82.7% and an estimated 78.3%. Omics data were used in 13.6% and an estimated 12.1%, free text in 7.4% and 9.6%. Oncology was the commonest specialty (37.0% and an estimated 38.7%), and the median primary model used two data sources in both corpora. Most studies analysed retrospective data (76.5% and an estimated 78.0%). The median test-set size was 216.5 in the legacy corpus and 150 in the update corpus (unweighted).

The multimodal model outperformed the strongest eligible unimodal comparator on the study's own primary metric in 85.2% of legacy studies (69 of 81) and an estimated 89.3% of update studies (85 of 96 unweighted). Pooled across periods, the figure was 88.6% (95% CI, 83.5–93.7). Eight studies did not report a numeric value for both models and count here as not showing a gain, so these figures are lower bounds. Among the 169 studies with both values, the multimodal model matched or fell short of its comparator in 15 (legacy 9 of 78; update 6 of 91).

AUROC was reported for both models in 127 studies (legacy 54, update 73). The pooled, design-weighted median gain was 0.042 (IQR, 0.020–0.100) (Fig. 1a and Extended Data Table 2). The legacy median was 0.033 (IQR, 0.010–0.061) and the estimated update median 0.046 (0.020–0.100). Mean gains in the two corpora differed by 0.008 (s.e. 0.013). An estimated 7.1% of these studies showed no gain in AUROC and 21.8% a gain of 0.10 or more. The right tail was long. The largest gain, 0.532, came from a proof-of-concept study of 18 patients evaluated by leave-one-out cross-validation, in which the image-only comparator had an AUROC of 0.26.

### *Gains were smaller when every data source was benchmarked*

We counted a study as completely benchmarked when every data source in its multimodal model also had its own separately fitted model, evaluated on the same test set. For a model that combines a CT scan with laboratory values and clinical history, that means an image-only, a laboratory-only and a clinical-only model. Studies that fell short left at least one constituent source without a model of its own. Nested designs, which add a new source to an existing model, usually fall in this group by construction.

Completely benchmarked studies reported smaller gains. We fitted a survey-weighted regression of ΔAUROC on complete benchmarking, external validation of the gain, log test-set size and study period to the 110 studies with AUROC for both models and a numeric test-set size. The coefficient for complete benchmarking was −0.030 (95% CI, −0.056 to −0.005; $P = 0.020$) (Fig. 2 and Extended Data Table 3). Without adjustment, the weighted mean gain was 0.046 in completely benchmarked studies (79 studies) and 0.080 in the rest (48 studies) (Fig. 1b and Supplementary Table S1).

The coefficient stayed between −0.028 and −0.035 when preprints, methods papers or studies of borderline multimodal status were excluded, separately or together (all $P < 0.05$; Extended Data Table 4). In post hoc checks it was similar in an unweighted model (−0.027; $P = 0.040$), after removing the 18-patient study (−0.026) and after winsorizing ΔAUROC at the 5th and 95th percentiles (−0.025), and it was negative in 98.4% of 2,000 weighted bootstrap resamples (Supplementary Tables S2, S3). Median regression gave −0.026 (95% CI, −0.077 to 0.026). On the logit scale, which has no ceiling, the coefficient was −0.168 (95% CI, −0.322 to −0.014; $P = 0.033$; post hoc). Adding the number of modality types (one versus two or more), the modality types and the prediction task gave −0.026 (95% CI, −0.050 to −0.003; $P = 0.030$; post hoc). Dropping test-set size from the model kept all 127 studies with AUROC for both models and gave −0.034 (95% CI, −0.057 to −0.012; $P = 0.003$; post hoc) (Supplementary Table S2).

Prediction methodology anticipates this direction. When a constituent source lacks its own model, the reference can be weaker than the best single-source model the data would have supported, and on the AUROC scale the room for a gain narrows as the reference improves[6–8]. Our data show that gradient across studies (exploratory; Supplementary Table S4).

The median gain was 0.127 when the comparator's AUROC was 0.70 or lower (16 studies), 0.046 between 0.70 and 0.80 (45 studies), 0.042 between 0.80 and 0.90 (48 studies) and 0.023 above 0.90 (18 studies).

Comparisons within studies point the same way (post hoc; Extended Data Fig. 1). In 78 completely benchmarked studies, we took the best single-source model for each constituent source and asked how the reported gain would have changed had one of these models been missing. Had the strongest source lacked a model of its own, the gain against the best remaining source would have been larger by a design-weighted mean of 0.062 (95% CI, 0.052–0.073), raising the average reported gain in these studies from 0.046 to 0.108. The increase would have been 0.05 or more in an estimated 47% of studies. Had only one source been modelled, chosen at random, the gain would have been larger by 0.035 (95% CI, 0.029–0.041), close to the difference between incompletely and completely benchmarked studies (0.080 versus 0.046). In 27 of the 60 studies with a structured clinical model, that model was the strongest single-source model or tied with it.

External validation of the gain was not associated with its size ($\beta$ = 0.009; 95% CI, −0.013 to 0.030; P = 0.43). Larger test sets went with smaller gains ($\beta$ = −0.0069 per unit of log test-set size, about 0.016 per tenfold increase; 95% CI, −0.0127 to −0.0012; P = 0.019).

### *Fewer recent studies benchmarked every data source*

Complete benchmarking was reported in 69.1% of legacy studies (56 of 81) and an estimated 46.0% of update studies (50 of 96 unweighted), a difference of −23.1 percentage points (95% CI, −38.0 to −8.3; P = 0.002) (Fig. 3). The decline was also present without weights (52.1%), within the screen-included stratum alone (55.4%) and among studies combining two data sources (65.2% legacy versus 47.9% update, unweighted) (Supplementary Tables S5, S6, S7).

Decision-utility analysis moved the other way, from 9.9% of legacy studies (8 of 81) to an estimated 24.1% of update studies (25 of 96 unweighted), a difference of 14.2 percentage points (95% CI, 2.9 to 25.5; P = 0.014). Reporting of uncertainty for the gain (46.9% and an estimated 41.2%), external validation of the gain (28.4% and 20.9%) and calibration (16.0% and 26.5%) showed no detectable change. Nor did the overall distribution of value claims (Rao–Scott P = 0.10), although the share claiming improved care rose (see below).

***Few studies tested the gain outside the development data***

Every eligible study contained a qualifying multimodal–unimodal comparison on the same outcome and evaluation set, and a numeric gain on the primary metric could be calculated for 169 of 177 studies. Beyond that point reporting fell away (Fig. 4 and Extended Data Table 5). An uncertainty estimate for the gain itself (a confidence interval, paired test or bootstrap of the difference between models) was reported in 46.9% of legacy studies and an estimated 41.2% of update studies. Both the multimodal model and its comparator were evaluated on data from a different site, population or setting in 28.4% and an estimated 20.9%. Another 4.9% and 4.0% evaluated both on a later period at the same site. Calibration of the multimodal model was assessed in 16.0% and an estimated 26.5%, and decision utility, by decision-curve analysis or another explicit threshold-based analysis, in 9.9% and an estimated 24.1%.

Few studies combined these. Pooled across periods, an estimated 12.1% of studies (95% CI, 7.8–16.4) reported a gain together with an uncertainty estimate and confirmed it on external data (Fig. 5). An estimated 21.8% (14.0–29.7) assessed decision utility, 9.2% did so in a study that also validated the gain externally, and 4.3% (1.7–7.0) combined decision utility with external validation and calibration. Read as a strict sequence, in which a study reaches a level only if it meets every earlier one, 12.3% of legacy studies and an estimated 6.4% of update studies reached external validation, and 2.5% and an estimated 3.2% reached decision utility (Fig. 4b).

Part of the drop in the strict sequence comes from how studies divide the work. Calibration and decision utility usually appeared together, and mostly in studies that had not benchmarked every source (exploratory; Supplementary Tables S8, S9 and Extended Data Fig. 2). An estimated 58% of studies reporting calibration also reported decision utility, and 66% of those reporting decision utility also reported calibration. Completely benchmarked studies reported calibration less often than the rest (an estimated 18% versus 32%) and decision utility less often (16% versus 28%). The 13 studies whose titles named a radiomics model or a nomogram reported calibration in 62% and decision curves in 69%, against 20% and 15% of other studies.

Two questions that decide whether an added source can be used in practice were rarely asked. The trained multimodal model was evaluated with an entire modality absent, as happens when a scan or assay has not been done, in 3.7% of legacy studies (3 of 81) and an estimated 7.2% of update studies (9 of 96 unweighted). The incremental burden or cost of acquiring the added source was quantified in one legacy study and two update studies (1.2% and an estimated 2.6%).

### *No eligible study evaluated a model prospectively*

None of the 81 legacy studies and none of the 96 update studies evaluated its model prospectively, whether as a prospective observational study, a silent deployment or an interventional trial (upper 95% confidence bounds, 4.5% and 3.8%). We counted an evaluation as prospective only when the model's predictions were generated prospectively. Seven legacy studies and nine update studies (unweighted) had recruited their participants prospectively, but none of them generated predictions that way. Evidence about clinical decisions came only from retrospective modelling, mainly decision curves in an estimated 22% of studies, and most studies offered none.

***Claims of clinical value went beyond the evidence shown***

We coded the strongest explicit claim of value in each study's abstract or conclusion (Fig. 6 and Extended Data Table 6). Claims confined to predictive performance were the exception, made by 16.0% of legacy studies and an estimated 12.3% of update studies. Most claimed potential clinical usefulness (80.2% and an estimated 74.0%), a category that includes hedged statements that the model may or could support clinical decisions. A minority stated that the model would improve care or patient outcomes: 3.7% of legacy studies (3 of 81) and an estimated 13.7% of update studies (12 of 96 unweighted; $P = 0.038$ for the difference).

Broad claims beyond predictive performance did not increase detectably with the amount of evidence reported. Neither decision utility nor prospective evaluation was reported by 90.1% of legacy studies (73 of 81) and an estimated 75.9% of update studies. Among these studies, 83.6% of legacy studies (61 of 73) and an estimated 85.9% of update studies (60 of 71 unweighted) still claimed potential clinical usefulness or improved care, almost the same as the 84.0% and estimated 87.7% among all studies. In an exploratory model, the odds of a claim beyond performance did not rise detectably with the number of core evidence items a study met (odds ratio per item, 1.17; 95% CI, 0.80–1.70).

Explicit claims of improved care were more frequent in studies that had run a decision-curve analysis (an estimated 32.2% versus 6.5% of other studies) or had validated the gain externally (24.0% versus 8.7%) (exploratory; Supplementary Table S10). None of these studies had evaluated its model prospectively. Net-benefit curves and external test sets appear to be read as evidence that care improves, although neither measures what happens to patients when clinicians use the model[11,12].

The qualifying comparison appeared in the abstract in 76.5% of legacy studies and an estimated 76.4% of update studies, and the gains reported in those studies were not larger than in studies that reported the comparison only in the full text (adjusted difference in ΔAUROC, −0.015; 95% CI, −0.053 to 0.022; $P = 0.42$).

## Discussion

Most studies of multimodal clinical AI show that adding a data source improves discrimination. Fewer compare the multimodal model with a separate model for every source. Tests of the gain on external data or at a decision threshold are uncommon, and none of the 177 studies evaluated a model in clinical use. For a clinician deciding whether to order the additional scan or assay a model requires, this literature shows that the added data carry information and says little about whether acquiring them improves decisions. Reviews of deep-learning studies in medical imaging found few external or prospective evaluations[15,16], and a review of machine-learning prediction models found frequent recommendations for clinical use without external validation[17]. Our results show the same pattern for multimodal models, with an added comparator problem particular to models that combine several sources.

A reported ΔAUROC compares two models, and its size depends on the strength of the comparator. When one constituent source has no model of its own, a reader cannot tell whether the multimodal model beats the best single-source option or only one of the options. In one benchmark of multi-omics integration, a single source or a small subset of sources matched or outperformed the full set in several datasets, so fewer tests would have sufficed[18]. Among the completely benchmarked studies in this sample, leaving out the strongest source's model would have more than doubled the average reported gain. In a study that already holds the data, fitting a model for every constituent source on the same test set adds little work, and printing the comparator's AUROC beside the gain lets readers judge the gain against its ceiling[6–8]. Reporting guidance for prediction-model studies sets out what to report about each model's development and performance[19]. For multimodal models, the set of comparators needs to be specified as well. Complete benchmarking fell between the two periods and was missing from an estimated 54.0% of recent studies.

A gain in discrimination is the opening step of the case for an added test[13,20]. The later steps concern decisions: whether predictions are calibrated in the population where the model will be used[9,21], whether acting on them at a clinically chosen threshold does more good than current practice[11], what the model does when the added source is missing[2,22], and what acquiring that source costs in money, time and patient burden[13,20]. Decision-curve analysis became more common between periods, but it stayed concentrated in a group of studies that often did not benchmark every source, and it was the analysis most often followed by a claim of improved care. Net benefit estimated on retrospective data shows whether a model could be useful at a given threshold[11,23]. Whether clinicians act on the model and whether patients benefit can only be learned prospectively[12,24,25], and we found no prospective evaluation in either period.

Claims of usefulness were close to universal, whatever evidence a study reported. Conclusions stronger than the evidence have been documented across biomedical research[26] and in studies of deep learning and machine-learning prediction models[16,17]. A statement that a model could support clinical decisions is hard to test. Journals could ask that claims of clinical usefulness name the evidence level the study reached, and that claims of improved care be kept for studies that measured care or outcomes.

A blinded second reviewer independently recoded a random sample of 36 studies (20%), with agreement of 88.6% to 100% on the evidence items and 94.3% (κ 0.89) on complete benchmarking.

Unweighted estimates, and estimates without stratum B, differed from the design-weighted prevalences by at most 6.3 percentage points (Extended Data Fig. 3). An audit of Embase estimated 73 further eligible studies outside the PubMed frame (95% CI, 50–100); adding them changed update-period prevalences by 4.5 percentage points or less, but the increase in claims of improved care was no longer statistically significant (Methods). Full text was unavailable for 2 legacy and 16 update records. Available and unavailable records did not differ detectably in publication year (P = 0.30 legacy, 0.11 update) or field (P = 0.41, 0.66).

Two changes would make these studies more useful to clinicians: a separately fitted model for every constituent data source, reported on the same test set, and prospective evaluation of clinical decisions before an added modality becomes part of routine care.

## Methods

### *Study design*

This was a meta-research study of multimodal clinical prediction models reported in the literature. It combined a census of the studies included in an earlier systematic review (the legacy corpus) with a stratified probability sample of records indexed since (the update corpus). The unit of analysis was a study, represented by one primary multimodal-versus-unimodal comparison. The study analysed published articles and eligible preprints and involved no human participants.

### *Legacy corpus*

The legacy corpus comprised the 97 studies included in the systematic review by Benani et al.[14], which searched PubMed/MEDLINE through 27 January 2025. PMIDs were available in the review's supplementary dataset for 57 studies. The remainder were matched by title, year and first author before full-text retrieval, and every record's identity was checked against the first page of its PDF. Legacy studies were analysed as a census with weight 1.

### *Update corpus and sampling*

We reran the search concepts of the earlier review in PubMed on 1 October 2026, for records published from 28 January 2025, the day after the earlier review's search date, to 30 September 2026. The complete search strategy is provided in the Supplementary Methods. The search returned 926 unique records.

Titles and abstracts were screened with a large language model (GPT-5.6, OpenAI) in iterative passes. A first pass classified records as include, unclear or exclude. Unclear records were re-reviewed under a forced include-or-exclude rule, followed by a separate adjudication pass with the same configuration. Because inclusion at this stage required a comparator visible in the abstract, and qualifying comparisons are often reported only in the full text, records excluded at screening were also sampled. The 926 records formed three strata: records included at screening (stratum I; N = 386), records included by the first binary pass but excluded at adjudication (stratum A; N = 153) and records excluded by both binary passes (stratum B; N = 387). We drew simple random samples of 120, 25 and 25 records within strata, each stratum sorted by PMID before drawing, with a fixed random seed. Because every stratum was sampled, screening errors reduce precision without biasing frame-level estimates.

Sampled records were matched to the legacy corpus by PMID, DOI and normalized title. Three sampled records were legacy studies and were analysed only in the legacy corpus. One sampled preprint had a peer-reviewed version that was also in the sample, and only the peer-reviewed version was analysed. For another sampled preprint, a peer-reviewed version had been published before coding, and that version was coded instead. Records whose full text could not be obtained (2 legacy, 16 update) were kept in the sample as nonrespondents and were never replaced.

### *Eligibility*

A study was eligible when, on full text, it met three criteria, applied identically to both corpora.

1. Multimodal model. The model used two or more independently acquired data sources. Features derived from the same raw data, sequences within one examination and the same acquisition type at different time points counted as one modality. Variables were classified by source, so measurements taken from images belonged to imaging.

2. Clinical scope. The model used person-level human data to predict or support a clinically meaningful outcome: a disease, injury or condition defined by clinical criteria, or a clinical decision, treatment or care-utilization outcome. Validated thresholds on clinical screening instruments qualified. Education, wellness or performance tasks and molecular-level predictions without person-level outcomes did not.

3. Qualifying comparator. The study reported a model built on a strict subset of the modalities, evaluated on the same data and outcome: a separately fitted unimodal model (including a clinical-only model refitted on the study data), a modality-withheld ablation or a nested addition. Architecture-only ablations, results from other studies, clinical scores applied as published and nominally clinical models that contained image-derived measurements did not qualify. A comparator anywhere in the full text, including figures and supplements, counted.

Reviews, perspectives, protocols and editorials were excluded. Methods or resource papers were eligible when evaluated on real clinical data for a clinical task, and were flagged. Preprints were included only when no peer-reviewed version existed at the time of coding, and were flagged.

### *Primary comparison*

Each study contributed one comparison: the authors' designated primary multimodal model against the strongest eligible unimodal comparator, on the same primary test set and primary metric. When authors designated no primary model, we took, in order, the multimodal model named in the abstract, the model proposed as final in the Methods, the all-modality model sharing the comparators' architecture, and the first-listed all-modality model in the main results table. The primary multimodal model was never selected by its score. With several outcomes or tasks, we used the designated primary outcome or, failing that, the first listed. The primary metric was the authors' designated metric, otherwise the metric of the main between-model comparison reported in the abstract, otherwise AUROC. Values reported only in figures were read from the figure and marked as approximate. When a paper reported inconsistent values, we used the main results table that contained the primary comparison.

***Coded items***

The evidence items were defined in a codebook (version 1.1) before coding of the analysed records began. Each was coded as met or not met:

- C1, the gain is quantifiable: both models reported for the same outcome, metric and evaluation set.
- C2, complete unimodal benchmarking: every constituent modality has its own separately fitted unimodal model on the same evaluation set. Masking inputs of the trained multimodal model did not count.
- C3, uncertainty of the gain: a confidence interval, paired test or bootstrap of the difference itself. Separate intervals for each model did not count.

- C4, external validation of the gain: both models evaluated on data from a structurally distinct site, population or setting[27]. Temporal validation at the same site was recorded but did not meet the item.
- C5, calibration of the multimodal model: a calibration plot, calibration intercept or slope, or an observed-versus-predicted comparison[21]. A Brier score alone did not count.
- C6, decision utility: decision-curve or net-benefit analysis[10,11], or an explicit threshold-based utility or cost–benefit analysis. Net reclassification, integrated discrimination improvement or sensitivity and specificity alone did not count[28].
- C7, performance with a modality absent: the trained multimodal model evaluated with an entire modality missing, by masking, dropout or a missing-modality evaluation.
- C8, acquisition burden: the incremental burden or cost of acquiring the added modality quantified (recorded as none, discussed or quantified; only quantified met the item). Computational cost did not count.
- C9, prospective evaluation: predictions generated prospectively in an observational evaluation, a silent deployment[29] or an interventional trial[12]. Prospective data collection alone did not count.

We also recorded the primary metric and the values of both models on it, the AUROC of both models when reported, test-set size and number of events, how the primary model was identified, and where the comparison was first reported (abstract or full text). ΔAUROC was defined as the multimodal model's AUROC minus the comparator's. Test-set size was the number of patients contributing predictions. For cross-validation, it was the number of patients contributing out-of-fold predictions. The value claim was the highest explicit claim in the abstract or conclusion, coded as performance only, potential clinical usefulness (including hedged statements that a model may or could help) or improved care or outcomes. Covariates were modality types, task, specialty, data source, model class, data-collection design, and preprint, methods-paper and borderline-modality flags.

***Coding procedure***

For every record, J.F. coded the eligibility criteria, evidence items, value claim and primary-comparison values from the full text, recording each value with its page, table, figure or supplement location and a short verbatim excerpt. For the legacy corpus, fields available in the earlier review's dataset (primary metric, comparator value, AUROC values, test-set size, modality types, task, specialty and model class) were carried over and checked, eligibility was re-applied under our criteria, and the multimodal value was checked against the full text, because the earlier review recorded the best-scoring multimodal model, whereas our rule used the designated one. In a post hoc comparison across 78 legacy studies, re-selecting the comparison by our rule did not systematically change the gain (median 0.043 with the earlier review's values, 0.038 with ours). With the earlier review's values, 92.3% of these 78 studies showed a gain, close to the 91% it reported[14]. Our rule gave 88.5%, and counting the three legacy studies without numeric values for both models as not showing a gain gives the 85.2% reported in the Results.

Operational clarifications (Supplementary Methods 3) were applied to all records. They covered figure-derived values, multiple outcomes, the provenance of image-derived variables, and the definitions of methods papers, data-collection design, modality types, test-set size, specialty and data source. None changed the evidence items or the primary-comparison rule.

A second reviewer independently recoded a stratified random sample of 36 studies (16 legacy and 20 update studies; fixed random seed) from the full text and codebook, without access to the first reviewer's values, evidence or notes. The second reviewer coded eligibility, the primary comparison and its values, every evidence item and the value claim, with page-level evidence for each judgment. Agreement was 97.2% for eligibility (35 of 36), 88.6% to 100% for the evidence items at the level used in the analysis (Cohen's κ 0.72 to 1.00 for the six items where it was estimable), including 94.3% for complete benchmarking (33 of 35; κ 0.89), and 85.7% for the value claim (κ 0.69). Disagreements were adjudicated against the full text. All analyses use the adjudicated values.

### *Statistical analysis*

Legacy proportions are census proportions with Wilson 95% confidence intervals[30]. Update estimates used a stratified design with strata I, A and B and finite-population corrections (survey package in R[31,32]). Records without full text were treated as nonrespondents within their stratum, giving design weights of $N_h/n_h$ among respondents: 3.48 (I), 7.65 (A) and 16.83 (B). Legacy duplicates and the superseded preprint were respondents outside the analysis domain. Eligible studies were analysed as a domain. Update intervals are Korn–Graubard intervals[33]. For estimates of exactly 0% or 100%, we used an exact Clopper–Pearson bound[34] on the unweighted domain size.

Pooled estimates and between-period contrasts combined the corpora in one design, with the legacy corpus as a separate stratum of weight 1 without a finite-population correction. Treating the legacy corpus as a finite census gave the same point estimates with narrower intervals. Differences in proportions were estimated as design-based contrasts with Wald tests, and the distribution of value claims was compared with a Rao–Scott second-order corrected F test[31,35].

Analyses of the gain used studies reporting AUROC for both models (127 studies). The primary model was a survey-weighted linear regression of ΔAUROC on complete unimodal benchmarking (C2), external validation of the gain (C4), the natural log of test-set size and period, among the 110 of these studies with a numeric test-set size. Sensitivity analyses repeated the prevalence estimates and the regression after excluding preprints (0 legacy, 2 update), methods papers (4, 8) and studies of borderline multimodal status (3, 7), separately and together (7, 15) (Extended Data Table 4 and Extended Data Fig. 4).

Post hoc analyses, labelled as such in the text, included alternative regression specifications (unweighted least squares, exclusion of influential observations or of the 18-patient study, winsorized, relative-gain, logit-scale and primary-metric outcomes, adjustment for the number and types of modalities and the task, omission of test-set size, and median regression[36] with bootstrap standard errors), a weighted bootstrap of the regression (2,000 resamples), composite combinations of evidence items, co-occurrence of evidence items and hierarchical clustering of evidence profiles, gains by comparator AUROC, and survey-weighted logistic regressions of value claims. Comparisons of available and unavailable full-text records used Wilcoxon tests for publication year and simulated Fisher's exact tests for a keyword-based field classification. Tests were two-sided with $\alpha = 0.05$, with no adjustment for multiplicity. P values from post hoc analyses are descriptive. Missing values were not imputed, and studies lacking an item were excluded from analyses that required it. Analyses used R 4.6.0 with survey 4.5 and quantreg 6.1 for median regression. Figures were drawn in Python 3.13 with Matplotlib 3.11.2[37].

### *Coverage audit of a second database*

We audited Embase to estimate how many eligible studies a second database would add to the update frame and whether they would change the estimates. The audit protocol was fixed before its sample was drawn, and the sensitivity analysis before any eligible study was coded. The search concepts were run in Embase on 7 October 2026 for the same publication dates, and 352 of the 1,182 unique records were full reports not linked to any study in the PubMed frame. A simple random sample of 100 of these was assessed under the same eligibility criteria, and J.F. coded the eligible studies with the same codebook and procedure. The 352 records were then treated as a fourth update stratum, with a design weight of 352/97 among the 97 sampled records with full text, and the prevalence estimates, the share of studies in which the multimodal model beat its strongest comparator, value claims and the primary regression were recomputed. Twenty sampled records were eligible additional studies, an estimated 73 (95% CI, 50–100) among the 352, equal to 16.7% of the eligible studies estimated for the PubMed frame. With them added, no update-period prevalence of an evidence item changed by more than 4.5 percentage points, and the coefficient for complete benchmarking was −0.041 (95% CI, −0.065 to −0.017). The increase in claims of improved care between periods and the coefficient for test-set size were no longer statistically significant (Supplementary Methods 4, Supplementary Tables S11–S14 and Supplementary Fig. 1).

### *Within-study comparator analysis*

This post hoc analysis was specified in writing before any value was extracted. It used the 79 analysed studies with complete unimodal benchmarking and AUROC for both models. For every constituent source of the primary multimodal model, J.F. extracted from the full text the AUROC of the best separately fitted single-source model on the primary outcome and test set, with its location and a verbatim excerpt. Sequences or phases of one examination counted as one source, models containing variables derived from another source were not used, and values shown only in figures were read from the figure. In every study, the highest single-source AUROC equalled the coded comparator value. With the per-source AUROCs ordered $U(1) \geq U(2) \geq \ldots$, the increase in the reported gain was $U(1) - U(2)$ had the strongest source not been modelled, $U(1)$ minus the mean of all single-source AUROCs had one source been modelled at random, and $U(1)$ minus the AUROC of the structured clinical model where one existed. Means and shares used the pooled design. One study was excluded because the AUROC of one source was reported only in an unavailable supplement (Supplementary Table S15).

### ***Data availability***

The coded dataset, with the location and excerpt supporting each judgment, is available from the corresponding author on reasonable request. The 97 legacy studies are listed in the supplementary dataset of the earlier review[14].

### ***Code availability***

Analysis code is available from the corresponding author on reasonable request.

### *Author contributions*

J.F. conceived and designed the study, developed the codebook, coded the studies, adjudicated coding disagreements, performed the statistical analysis and wrote the manuscript.

### *Competing interests*

The author declares no competing interests.

## Figure legends

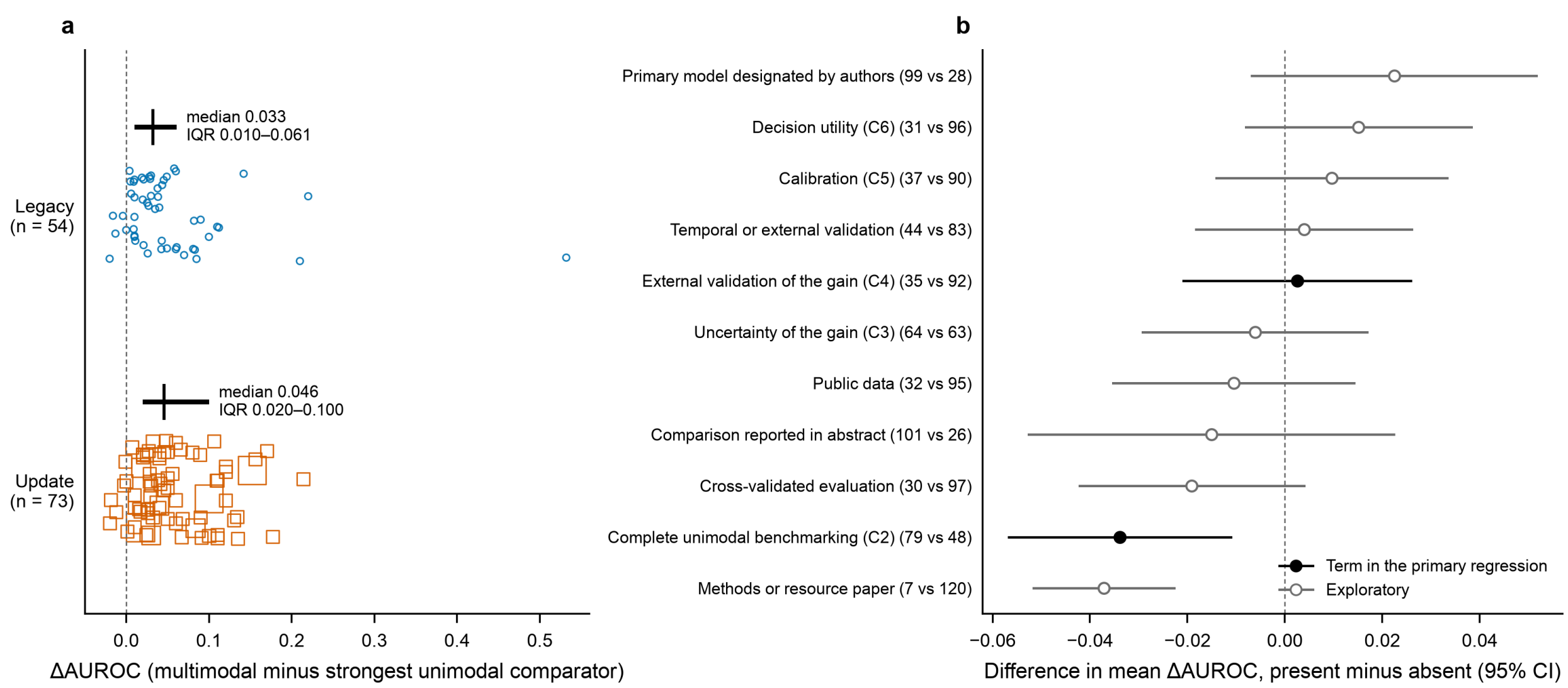

**Fig. 1 | Reported gain in AUROC. a**, Distribution of ΔAUROC (primary multimodal model minus strongest eligible unimodal comparator) in 127 studies reporting AUROC for both models (legacy 54, update 73). Each marker is one study; marker area is proportional to the design weight. Black bars mark the median (vertical tick) and IQR, design-weighted for the update corpus. **b**, Difference in mean ΔAUROC between studies with and without each feature, pooled and design-weighted, with 95% CIs. Filled markers, terms in the primary regression; open markers, exploratory.

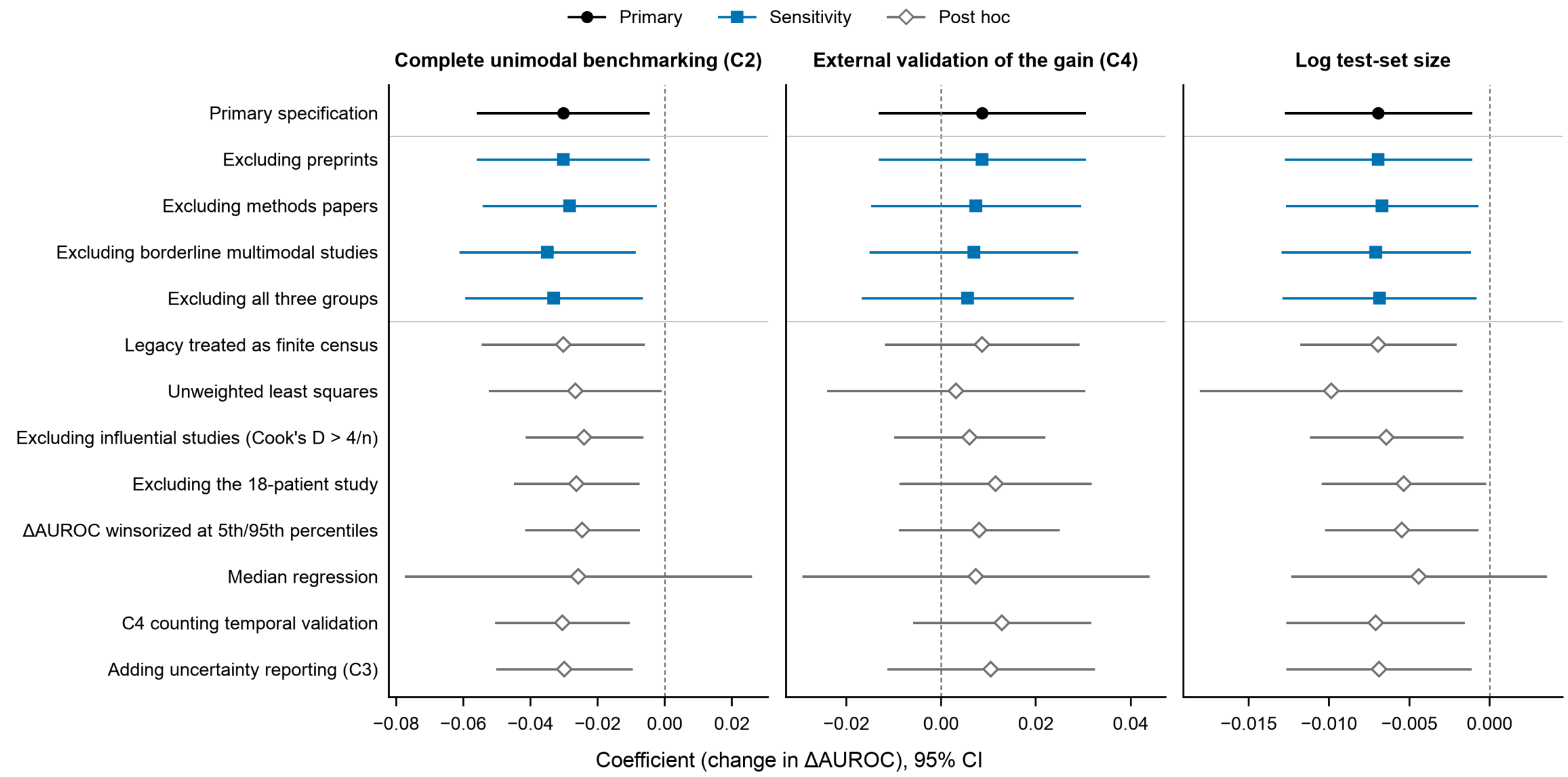

**Fig. 2 | Complete unimodal benchmarking and the size of the gain.** Coefficients (95% CI) from survey-weighted linear regression of ΔAUROC on complete benchmarking, external validation of the gain, log test-set size and period (n = 110), under the primary specification, sensitivity exclusions and post hoc robustness checks. Models with outcomes on other scales (relative gain, logit, primary metric), the adjusted model and the model without test-set size are in Supplementary Table S2.

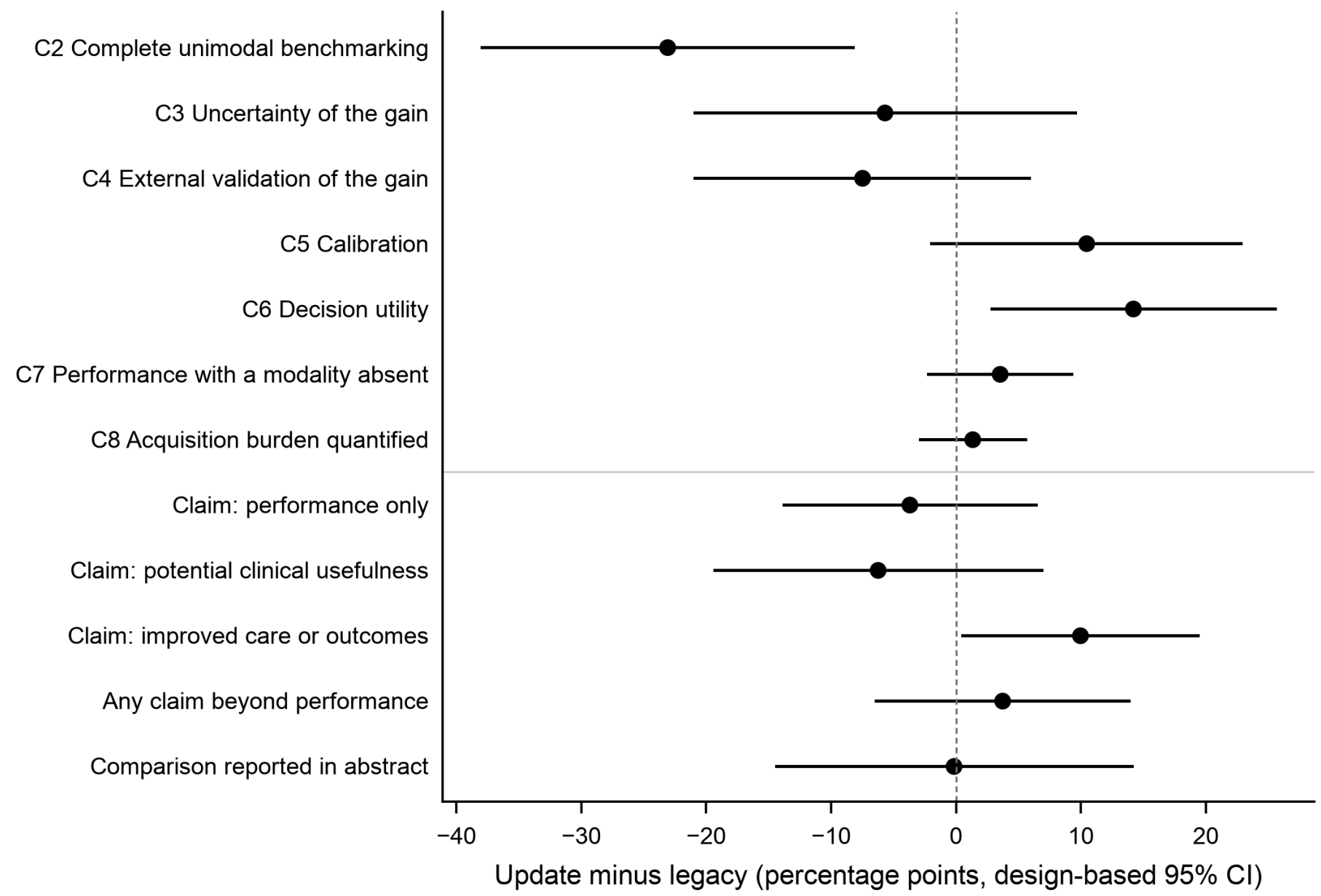

**Fig. 3 | Change between the legacy corpus and the update period.** Update minus legacy, in percentage points, for each evidence item, value claim and abstract visibility, with design-based 95% CIs.

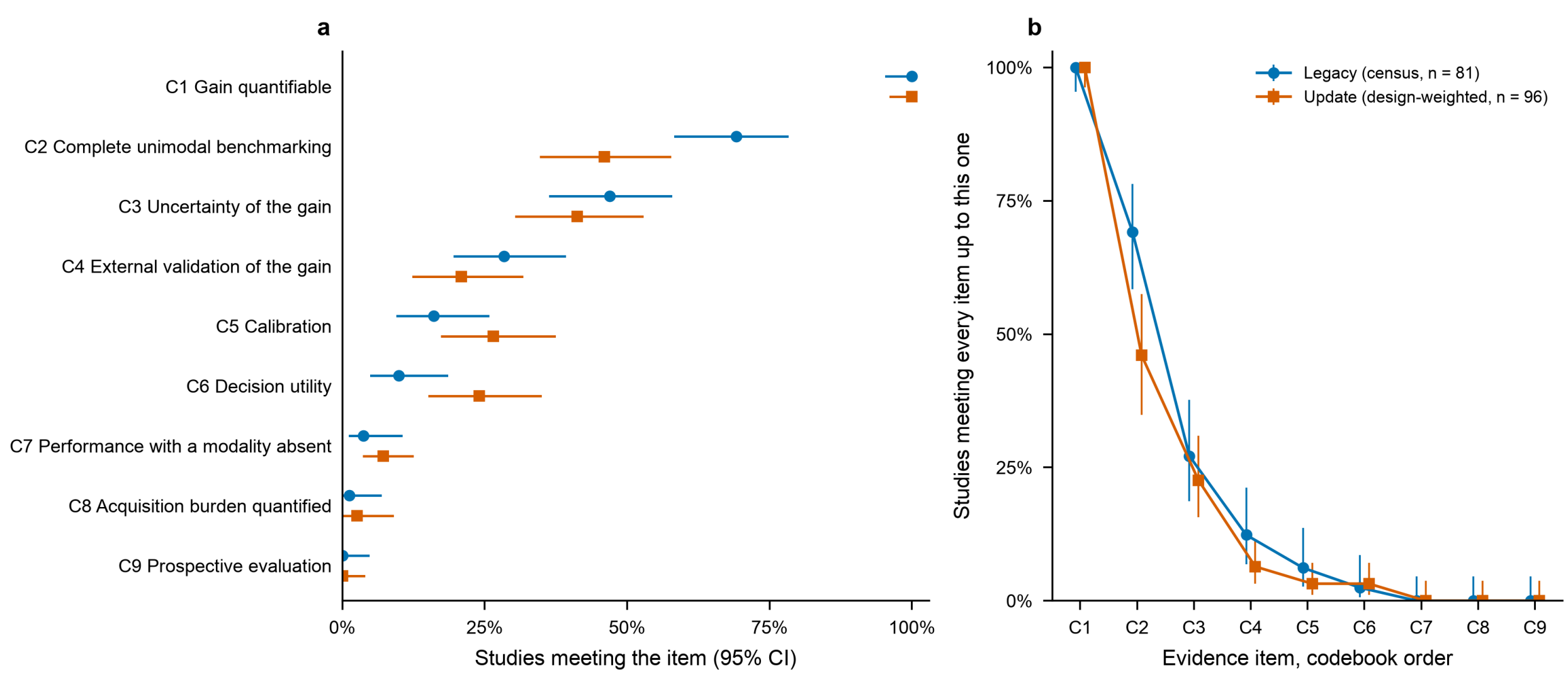

**Fig. 4 | Evidence reported beyond the gain. a**, Share of studies meeting each item (legacy: census of 81, Wilson 95% CI; update: 96 studies, design-weighted, Korn–Graubard 95% CI). **b**, Share meeting every item up to and including each one, in codebook order.

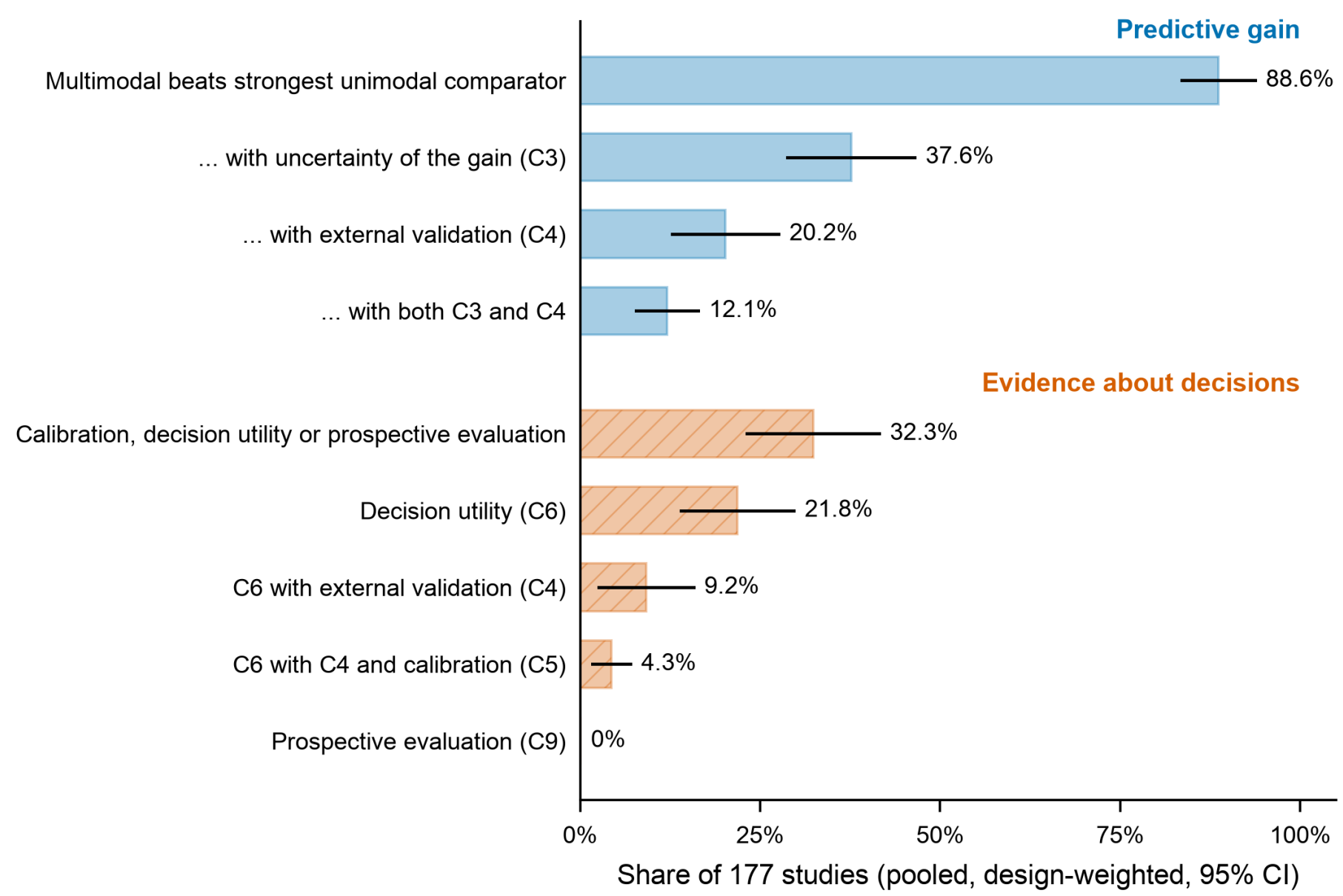

**Fig. 5 | From a reported gain to evidence about clinical decisions.** Pooled, design-weighted shares of 177 studies meeting combinations of evidence items, with 95% CIs. Combinations were defined after coding. Studies without a numeric value for both models on the primary metric count as not showing a gain.

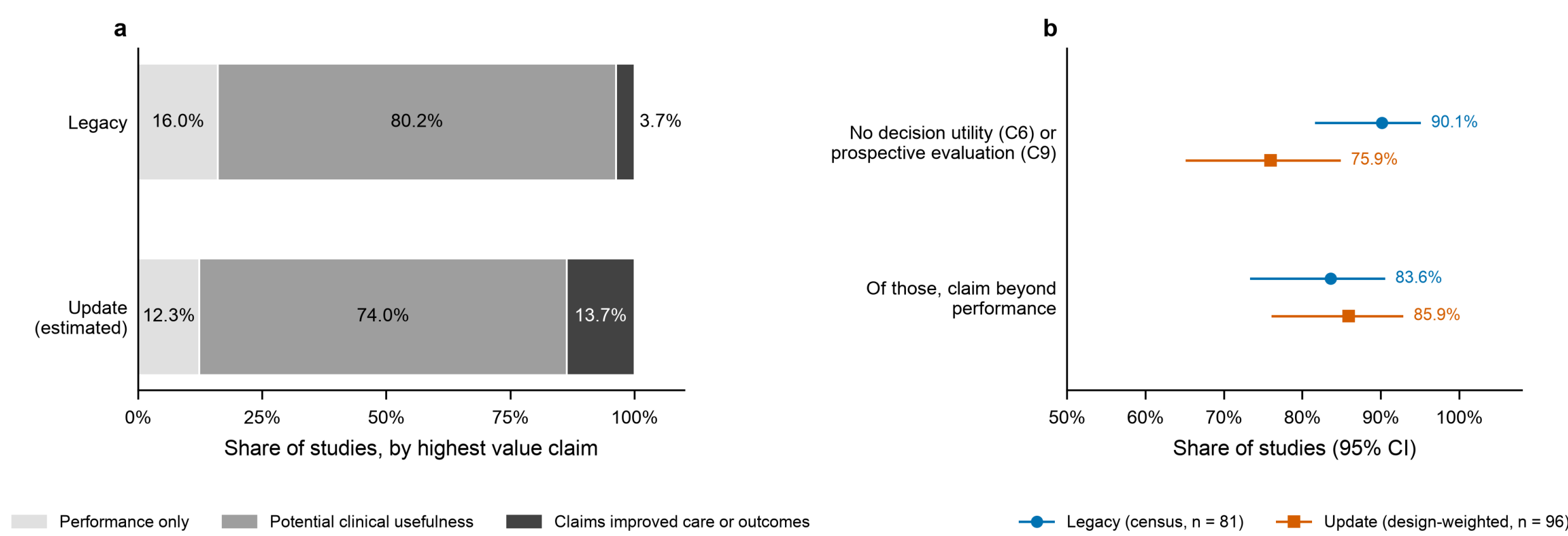

**Fig. 6 | Value claims and the evidence beneath them. a**, Highest explicit value claim by corpus. **b**, Share of studies with neither decision utility nor prospective evaluation, and the share of those claiming potential clinical usefulness or improved care, with 95% CIs. Update estimates are design-weighted.

# Extended Data

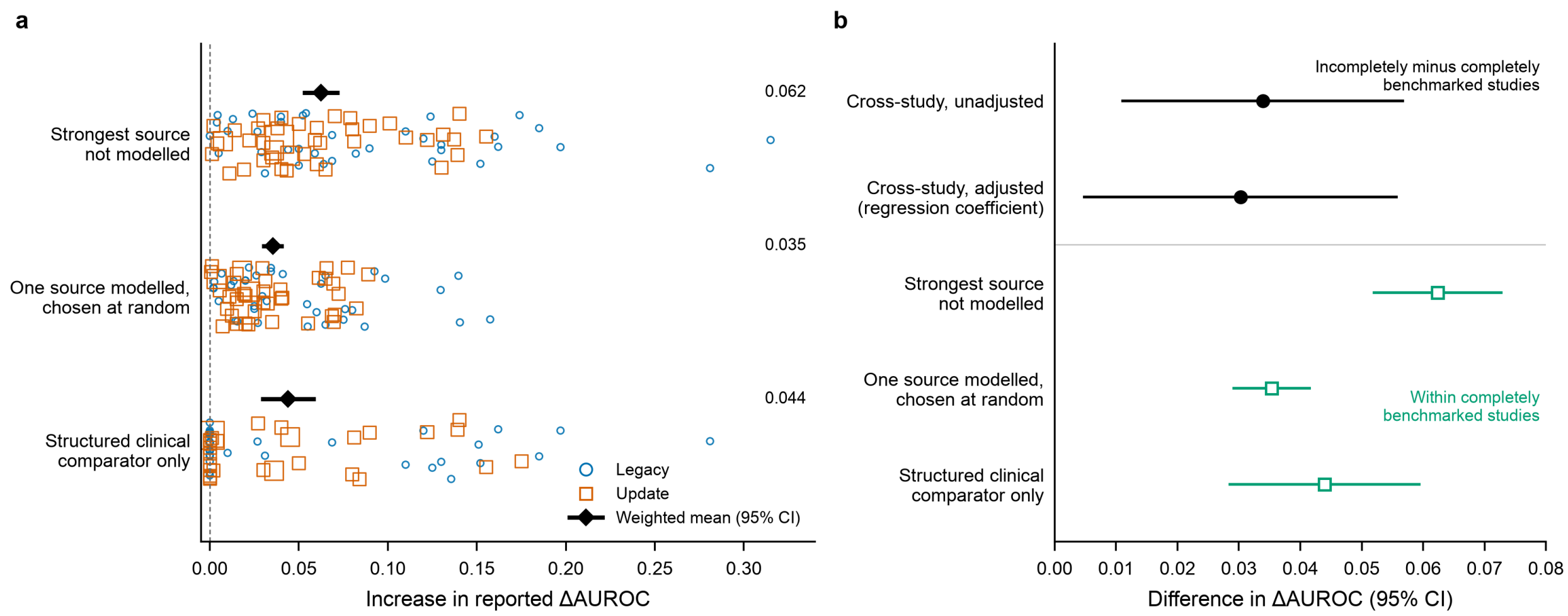

**Extended Data Fig. 1 | Gain against different single-source comparators within studies. a**, Increase in the reported ΔAUROC in 78 completely benchmarked studies had the strongest source not been modelled, had one source been modelled at random, or had only the structured clinical model been fitted (60 studies with such a model). Each marker is one study, with area proportional to the design weight; diamonds and bars show the design-weighted mean and 95% CI. **b**, The within-study increases beside the cross-study difference in ΔAUROC between incompletely and completely benchmarked studies, unadjusted and as the coefficient of the primary regression with its sign reversed, with 95% CIs. Post hoc.

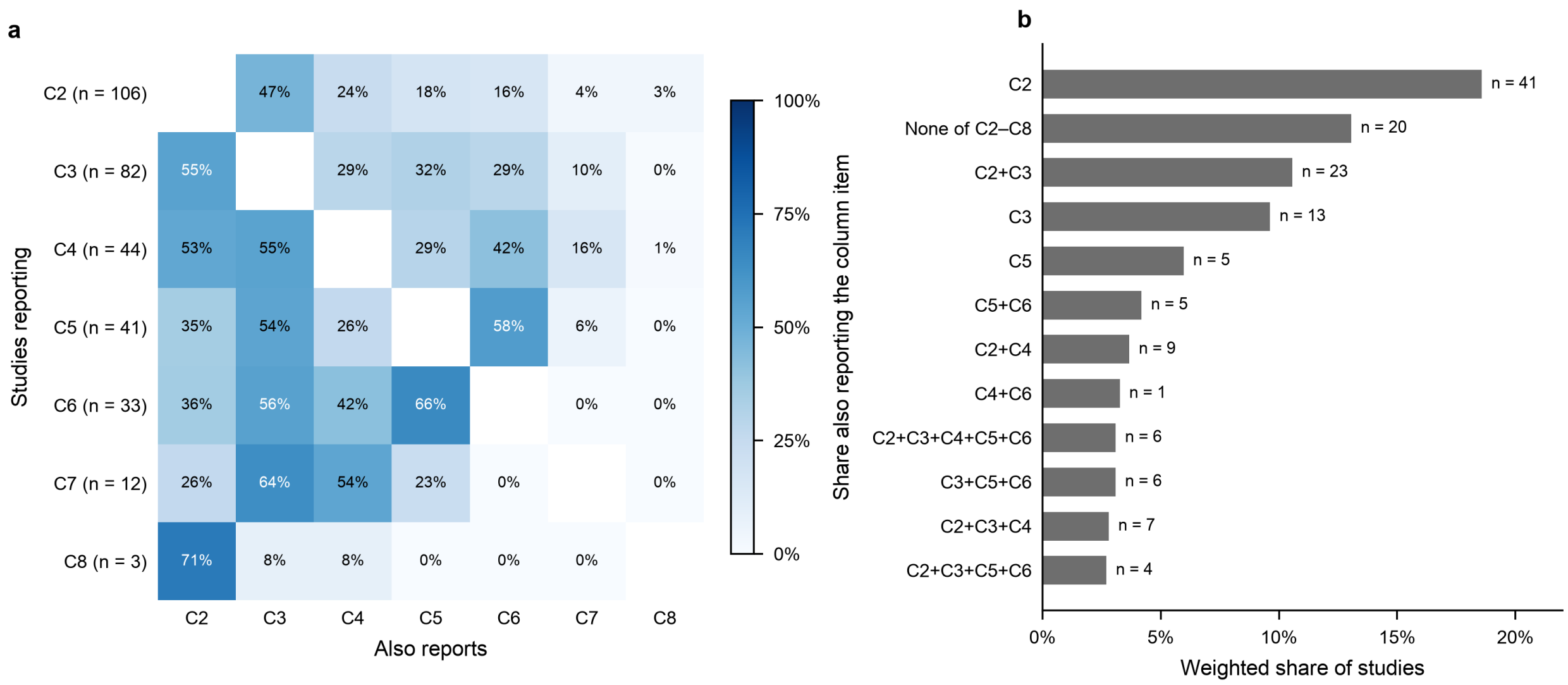

**Extended Data Fig. 2 | Structure of the evidence reported. a**, Share of studies reporting the row item that also report the column item. **b**, The 12 commonest combinations of items C2–C8; bar labels give unweighted counts. Pooled, design-weighted, exploratory.

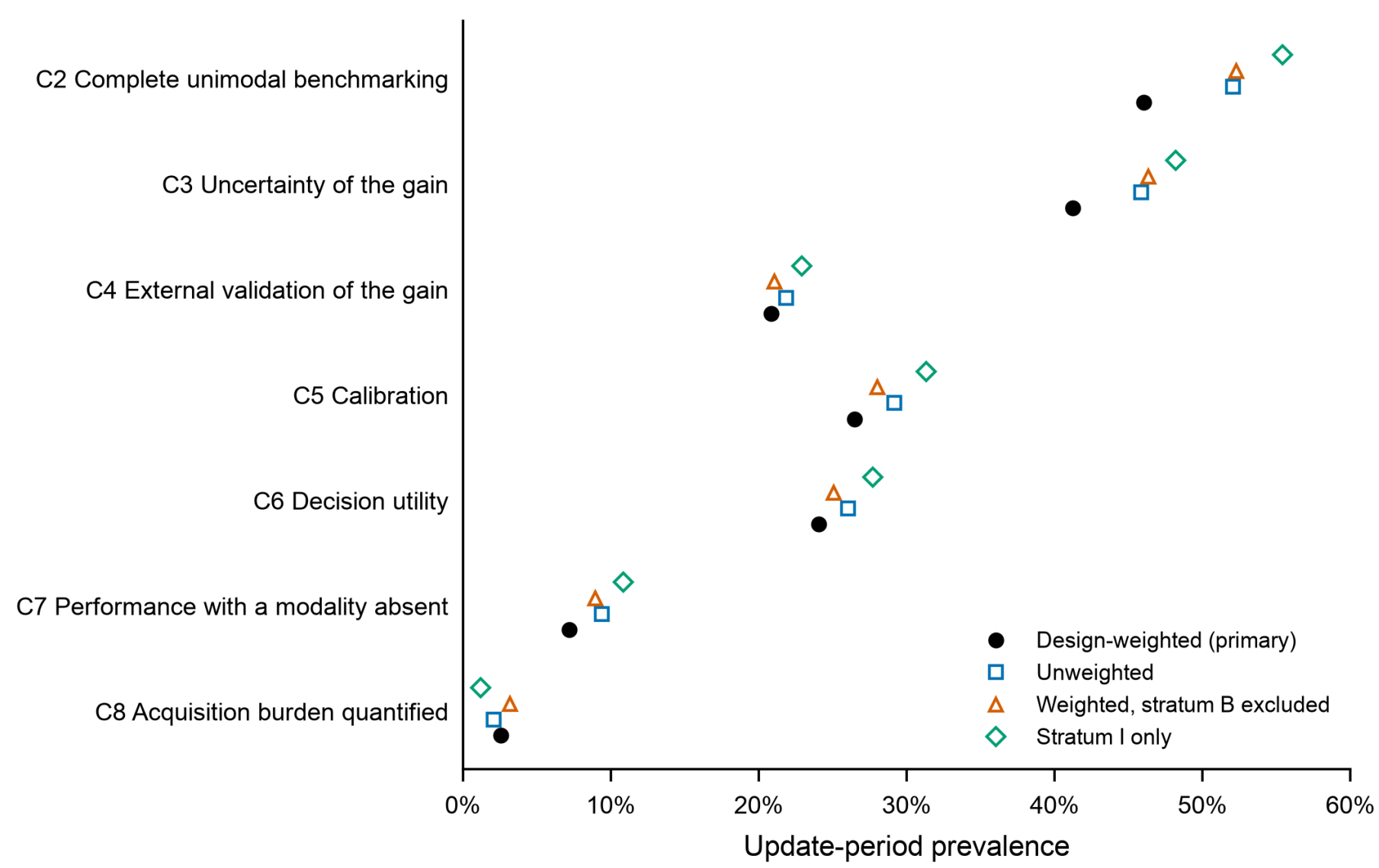

**Extended Data Fig. 3 | Update-period prevalence under alternative weighting.** Design weights (primary), no weights, design weights excluding stratum B, and stratum I alone.

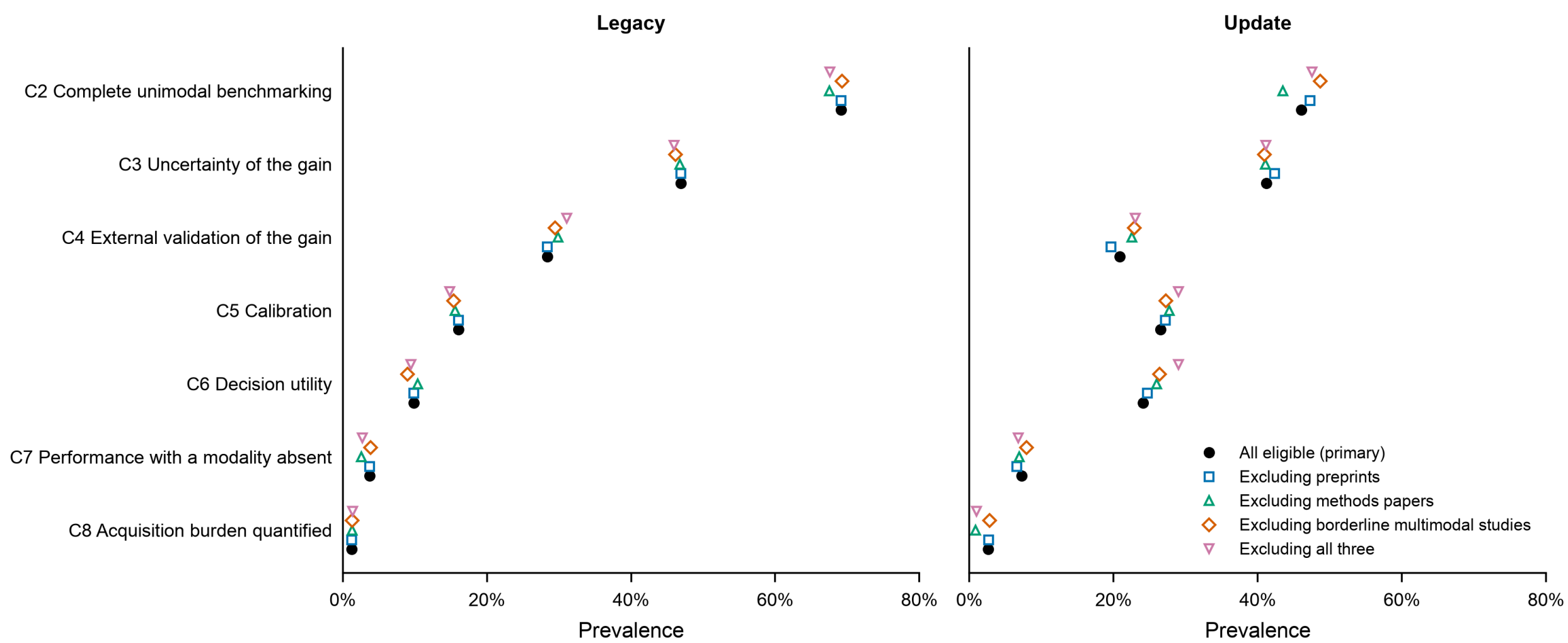

**Extended Data Fig. 4 | Prevalence of evidence items under the sensitivity exclusions.** Exclusion of preprints, methods papers and studies of borderline multimodal status, separately and together, by corpus.

## Extended Data Table 1 | Characteristics of eligible studies.

| Characteristic | Level | Legacy, n (%) | Update, n (unweighted) | Update, weighted % (95% CI) |
|---|---|---|---|---|
| Task | Diagnosis-detection | 39 (48.1) | 47 | 42.5 (32.3-53.1) |
| Task | Other clinical | 1 (1.2) | 2 | 1.6 (0.3-4.8) |
| Task | Prognosis-risk | 34 (42.0) | 33 | 39.7 (28.3-51.9) |
| Task | Treatment response | 7 (8.6) | 14 | 16.2 (8.3-27.4) |
| Specialty | Cardiovascular | 10 (12.3) | 10 | 13.0 (5.6-24.5) |
| Specialty | Neurology-psychiatry | 19 (23.5) | 14 | 15.3 (7.7-26.1) |
| Specialty | Oncology | 30 (37.0) | 37 | 38.7 (27.6-50.7) |
| Specialty | Ophthalmology | 5 (6.2) | 9 | 11.2 (4.4-22.5) |
| Specialty | Other | 11 (13.6) | 19 | 15.2 (9.9-21.9) |
| Specialty | Respiratory-critical care | 6 (7.4) | 7 | 6.6 (2.8-12.8) |
| Data source | Both | 3 (3.7) | 3 | 2.4 (0.7-6.0) |
| Data source | Institutional | 57 (70.4) | 66 | 71.0 (59.9-80.4) |
| Data source | Public | 21 (25.9) | 27 | 26.6 (17.4-37.7) |
| Model class | Classical ML | 11 (13.6) | 21 | 24.9 (14.9-37.4) |
| Model class | Deep learning | 67 (82.7) | 71 | 70.9 (58.5-81.4) |
| Model class | Foundation model or LLM | 3 (3.7) | 4 | 4.2 (1.2-10.2) |
| Data design | Mixed | 2 (2.5) | 2 | 1.6 (0.3-4.8) |
| Data design | Not reported | 10 (12.3) | 5 | 7.1 (1.6-18.8) |
| Data design | Prospective | 7 (8.6) | 9 | 13.4 (5.2-26.6) |
| Data design | Retrospective | 62 (76.5) | 80 | 78.0 (64.6-88.1) |
| Primary metric | Accuracy | 17 (21.0) | 8 | 7.4 (3.4-13.7) |
| Primary metric | AUPRC | 1 (1.2) | 3 | 2.4 (0.7-6.0) |
| Primary metric | AUROC | 52 (64.2) | 73 | 72.5 (60.0-82.9) |
| Primary metric | C-index | 8 (9.9) | 6 | 7.9 (2.1-19.3) |
| Primary metric | F1 | 2 (2.5) | 3 | 3.4 (0.7-9.4) |
| Primary metric | Other | 1 (1.2) | 3 | 6.4 (1.1-19.4) |
| Validation of the gain (C4) | External | 23 (28.4) | 21 | 20.9 (12.5-31.6) |
| Validation of the gain (C4) | None | 54 (66.7) | 70 | 75.1 (64.4-84.0) |
| Validation of the gain (C4) | Temporal | 4 (4.9) | 5 | 4.0 (1.6-8.2) |
| Modality type (primary model) | Imaging | 72 (88.9) | 84 | 85.4 (74.1-93.1) |
| Modality type (primary model) | Physiological signal | 8 (9.9) | 9 | 11.2 (4.4-22.5) |
| Modality type (primary model) | Structured clinical/EHR | 67 (82.7) | 74 | 78.3 (67.6-86.8) |
| Modality type (primary model) | Free text | 6 (7.4) | 12 | 9.6 (5.5-15.3) |
| Modality type (primary model) | Omics | 11 (13.6) | 10 | 12.1 (5.0-23.2) |
| Modality type (primary model) | Audio/video | 3 (3.7) | 1 | 0.8 (0.0-3.6) |
| Modality type (primary model) | Other | 0 (0.0) | 1 | 1.8 (0.1-8.6) |
| Flags | Preprint | 0 (0.0) | 2 | 2.6 (0.3-8.8) |
| Flags | Methods or resource paper | 4 (4.9) | 8 | 7.4 (3.4-13.7) |
| Flags | Borderline multimodal status | 3 (3.7) | 7 | 8.7 (2.7-19.8) |
| Flags | Comparison reported in abstract | 62 (76.5) | 79 | 76.4 (63.6-86.4) |

Legacy: census of eligible studies (Wilson 95% CI). Update: stratified design (strata I/A/B, FPC N = 386/153/387), full-text unavailable records as within-stratum nonresponse; eligible studies analysed as a domain; Korn-Graubard 95% CI. Modality types refer to the primary multimodal model selected by the primary-comparison rule (Methods). Median test-set size: legacy 216.5, update 150.

# Extended Data Table 2 | Reported gain in AUROC.

| Group | n | Mean (95% CI) | Median [IQR] | Range | <= 0, % | < 0.02, % | >= 0.05, % | >= 0.10, % |
|---|---|---|---|---|---|---|---|---|
| Legacy (census) | 54 | 0.054 (0.031 to 0.076) | 0.033 [0.010-0.061] | -0.020 to 0.532 | 9.3 | 31.5 | 31.5 | 13.0 |
| Update (weighted) | 73 | 0.062 (0.048 to 0.076) | 0.046 [0.020-0.100] | -0.020 to 0.214 | 6.7 | 23.6 | 45.8 | 23.3 |
| Pooled (Legacy weight 1 + Update weighted) | 127 | 0.061 (0.048 to 0.073) | 0.042 [0.020-0.100] | -0.020 to 0.532 | 7.1 | 24.7 | 43.7 | 21.8 |
| Legacy, abstract-visible | 40 | 0.056 (0.028 to 0.084) | 0.038 [0.021-0.060] | - | - | - | - | - |
| Update (weighted), abstract-visible | 61 | 0.057 (0.045 to 0.070) | 0.041 [0.020-0.100] | - | - | - | - | - |
| Pooled, abstract-visible | 101 | 0.057 (0.046 to 0.068) | 0.041 [0.020-0.090] | - | - | - | - | - |
| Legacy, full-text only | 14 | 0.048 (0.010 to 0.085) | 0.010 [0.004-0.096] | - | - | - | - | - |
| Update (weighted), full-text only | 12 | 0.077 (0.037 to 0.118) | 0.084 [0.027-0.120] | - | - | - | - | - |
| Pooled, full-text only | 26 | 0.072 (0.037 to 0.108) | 0.060 [0.024-0.120] | - | - | - | - | - |

Studies reporting AUROC for both the primary multimodal model and the strongest eligible unimodal comparator (legacy 54 of 81; update 73 of 96). Update and pooled estimates design-weighted; pooled = Legacy weight 1 plus Update weighted. Abstract-visibility rows: secondary analysis.

# Extended Data Table 3 | Survey-weighted regression of ΔAUROC.

| Specification | Term | Coefficient (95% CI) | SE | P | n |
|---|---|---|---|---|---|
| Primary (Legacy weight 1, no FPC; Update weighted with FPC) | Intercept | 0.1056 (0.0486 to 0.1625) | 0.0287 | <0.001 | 110 |
| Primary (Legacy weight 1, no FPC; Update weighted with FPC) | C4, external validation of the gain | 0.0086 (-0.0130 to 0.0302) | 0.0109 | 0.431 | 110 |
| Primary (Legacy weight 1, no FPC; Update weighted with FPC) | C2, complete unimodal benchmarking | -0.0303 (-0.0557 to -0.0049) | 0.0128 | 0.020 | 110 |
| Primary (Legacy weight 1, no FPC; Update weighted with FPC) | Log test-set size | -0.0069 (-0.0127 to -0.0012) | 0.0029 | 0.019 | 110 |
| Primary (Legacy weight 1, no FPC; Update weighted with FPC) | Corpus: Update (vs Legacy) | 0.0126 (-0.0134 to 0.0387) | 0.0131 | 0.339 | 110 |
| Variant: Legacy treated as census (FPC = 1) | Intercept | 0.1056 (0.0704 to 0.1407) | 0.0177 | <0.001 | 110 |
| Variant: Legacy treated as census (FPC = 1) | C4, external validation of the gain | 0.0086 (-0.0117 to 0.0289) | 0.0102 | 0.401 | 110 |
| Variant: Legacy treated as census (FPC = 1) | C2, complete unimodal benchmarking | -0.0303 (-0.0542 to -0.0063) | 0.0121 | 0.014 | 110 |
| Variant: Legacy treated as census (FPC = 1) | Log test-set size | -0.0069 (-0.0117 to -0.0021) | 0.0024 | 0.005 | 110 |
| Variant: Legacy treated as census (FPC = 1) | Corpus: Update (vs Legacy) | 0.0126 (0.0017 to 0.0236) | 0.0055 | 0.024 | 110 |

Model: ΔAUROC ~ C4 + C2 + log(test-set size) + corpus, among studies with AUROC for both models and a numeric test-set size. Two-sided tests, alpha 0.05. Variant row is a post hoc variance specification (Legacy treated as a finite census).

# Extended Data Table 4 | Sensitivity analyses for prevalences and for the regression.

**a**

| Exclusion set | Evidence item | Legacy % (n/N) | Update weighted % (95% CI) |
|---|---|---|---|
| None (primary) | C1 Gain quantifiable | 100.0 (81/81) | 100.0 (96.2-100.0) |
| None (primary) | C2 Complete unimodal benchmarking | 69.1 (56/81) | 46.0 (34.9-57.5) |
| None (primary) | C3 Uncertainty of the gain | 46.9 (38/81) | 41.2 (30.5-52.6) |
| None (primary) | C4 External validation of the gain | 28.4 (23/81) | 20.9 (12.5-31.6) |
| None (primary) | C5 Calibration | 16.0 (13/81) | 26.5 (17.5-37.2) |
| None (primary) | C6 Decision utility | 9.9 (8/81) | 24.1 (15.3-34.8) |
| None (primary) | C7 Performance with a modality absent | 3.7 (3/81) | 7.2 (3.8-12.3) |
| None (primary) | C8 Acquisition burden quantified | 1.2 (1/81) | 2.6 (0.3-8.8) |
| None (primary) | C9 Prospective evaluation | 0.0 (0/81) | 0.0 (0.0-3.8) |
| Exclude preprints | C1 Gain quantifiable | 100.0 (81/81) | 100.0 (96.2-100.0) |
| Exclude preprints | C2 Complete unimodal benchmarking | 69.1 (56/81) | 47.2 (35.8-58.9) |
| Exclude preprints | C3 Uncertainty of the gain | 46.9 (38/81) | 42.3 (31.3-53.9) |
| Exclude preprints | C4 External validation of the gain | 28.4 (23/81) | 19.6 (11.5-30.1) |
| Exclude preprints | C5 Calibration | 16.0 (13/81) | 27.2 (17.9-38.1) |
| Exclude preprints | C6 Decision utility | 9.9 (8/81) | 24.7 (15.7-35.6) |
| Exclude preprints | C7 Performance with a modality absent | 3.7 (3/81) | 6.6 (3.3-11.6) |
| Exclude preprints | C8 Acquisition burden quantified | 1.2 (1/81) | 2.6 (0.3-9.0) |
| Exclude preprints | C9 Prospective evaluation | 0.0 (0/81) | 0.0 (0.0-3.8) |
| Exclude methods papers | C1 Gain quantifiable | 100.0 (77/77) | 100.0 (95.9-100.0) |
| Exclude methods papers | C2 Complete unimodal benchmarking | 67.5 (52/77) | 43.5 (32.0-55.4) |
| Exclude methods papers | C3 Uncertainty of the gain | 46.8 (36/77) | 41.0 (29.7-53.1) |
| Exclude methods papers | C4 External validation of the gain | 29.9 (23/77) | 22.5 (13.5-34.0) |
| Exclude methods papers | C5 Calibration | 15.6 (12/77) | 27.7 (18.1-39.2) |
| Exclude methods papers | C6 Decision utility | 10.4 (8/77) | 26.0 (16.5-37.4) |
| Exclude methods papers | C7 Performance with a modality absent | 2.6 (2/77) | 6.9 (3.4-12.2) |
| Exclude methods papers | C8 Acquisition burden quantified | 1.3 (1/77) | 0.9 (0.1-3.8) |
| Exclude methods papers | C9 Prospective evaluation | 0.0 (0/77) | 0.0 (0.0-4.1) |
| Exclude borderline multimodal studies | C1 Gain quantifiable | 100.0 (78/78) | 100.0 (95.9-100.0) |
| Exclude borderline multimodal studies | C2 Complete unimodal benchmarking | 69.2 (54/78) | 48.6 (36.9-60.6) |
| Exclude borderline multimodal studies | C3 Uncertainty of the gain | 46.2 (36/78) | 40.9 (30.6-51.8) |
| Exclude borderline multimodal studies | C4 External validation of the gain | 29.5 (23/78) | 22.9 (13.8-34.2) |
| Exclude borderline multimodal studies | C5 Calibration | 15.4 (12/78) | 27.2 (17.7-38.6) |
| Exclude borderline multimodal studies | C6 Decision utility | 9.0 (7/78) | 26.4 (16.9-37.7) |
| Exclude borderline multimodal studies | C7 Performance with a modality absent | 3.8 (3/78) | 7.9 (4.1-13.5) |
| Exclude borderline multimodal studies | C8 Acquisition burden quantified | 1.3 (1/78) | 2.8 (0.4-9.6) |
| Exclude borderline multimodal studies | C9 Prospective evaluation | 0.0 (0/78) | 0.0 (0.0-4.1) |
| Exclude all three | C1 Gain quantifiable | 100.0 (74/74) | 100.0 (95.5-100.0) |
| Exclude all three | C2 Complete unimodal benchmarking | 67.6 (50/74) | 47.5 (35.2-60.2) |
| Exclude all three | C3 Uncertainty of the gain | 45.9 (34/74) | 41.1 (30.2-52.7) |
| Exclude all three | C4 External validation of the gain | 31.1 (23/74) | 23.0 (13.6-34.9) |
| Exclude all three | C5 Calibration | 14.9 (11/74) | 29.0 (18.6-41.3) |
| Exclude all three | C6 Decision utility | 9.5 (7/74) | 29.0 (18.6-41.3) |
| Exclude all three | C7 Performance with a modality | 2.7 (2/74) | 6.8 (3.2-12.4) |
| | absent | | |
| Exclude all three | C8 Acquisition burden quantified | 1.4 (1/74) | 1.0 (0.1-4.3) |
| Exclude all three | C9 Prospective evaluation | 0.0 (0/74) | 0.0 (0.0-4.5) |

Exclusions: preprints, methods papers and studies of borderline multimodal status. Studies excluded (legacy/update): preprints 0/2, methods 4/8, borderline 3/7, any 7/15

**b**

| Exclusion set | n | C4 coefficient (95% CI) | C2 coefficient (95% CI) | Log test-set size coefficient (95% CI) | C4 P | C2 P | Log test-set size P |
|---|---|---|---|---|---|---|---|
| None (primary) | 110 | 0.0086 (-0.0130 to 0.0302) | -0.0303 (-0.0557 to -0.0049) | -0.0069 (-0.0127 to -0.0012) | 0.431 | 0.020 | 0.019 |
| Exclude preprints | 110 | 0.0086 (-0.0130 to 0.0302) | -0.0303 (-0.0557 to -0.0049) | -0.0069 (-0.0127 to -0.0012) | 0.431 | 0.020 | 0.019 |
| Exclude methods papers | 106 | 0.0073 (-0.0146 to 0.0292) | -0.0284 (-0.0540 to -0.0028) | -0.0067 (-0.0126 to -0.0008) | 0.511 | 0.030 | 0.027 |
| Exclude borderline multimodal studies | 105 | 0.0069 (-0.0149 to 0.0287) | -0.0349 (-0.0609 to -0.0090) | -0.0071 (-0.0129 to -0.0013) | 0.533 | 0.009 | 0.018 |
| Exclude all three | 101 | 0.0055 (-0.0166 to 0.0277) | -0.0330 (-0.0591 to -0.0069) | -0.0069 (-0.0128 to -0.0009) | 0.620 | 0.014 | 0.025 |

Survey-weighted regression as in Extended Data Table 3.

# Extended Data Table 5 | Evidence items: marginal and cumulative prevalence, and legacy–update differences.

| Evidence item | Legacy marginal, n/N | Legacy marginal, % (95% CI) | Update marginal, n/N (unweighted) | Update marginal, weighted % (95% CI) | Difference, pp (95% CI) | P | Legacy cumulative, % | Update cumulative, weighted % |
|---|---|---|---|---|---|---|---|---|
| C1 Gain quantifiable | 81/81 | 100.0 (95.5-100.0) | 96/96 | 100.0 (96.2-100.0) | - | - | 100.0 (95.5-100.0) | 100.0 (96.2-100.0) |
| C2 Complete unimodal benchmarking | 56/81 | 69.1 (58.4-78.1) | 50/96 | 46.0 (34.9-57.5) | -23.1 (-38.0 to -8.3) | 0.002 | 69.1 (58.4-78.1) | 46.0 (34.9-57.5) |
| C3 Uncertainty of the gain | 38/81 | 46.9 (36.4-57.7) | 44/96 | 41.2 (30.5-52.6) | -5.7 (-20.9 to +9.5) | 0.464 | 27.2 (18.7-37.7) | 22.6 (15.6-30.9) |
| C4 External validation of the gain | 23/81 | 28.4 (19.7-39.0) | 21/96 | 20.9 (12.5-31.6) | -7.5 (-20.9 to +5.8) | 0.270 | 12.3 (6.8-21.3) | 6.4 (3.2-11.3) |
| C5 Calibration | 13/81 | 16.0 (9.6-25.5) | 28/96 | 26.5 (17.5-37.2) | +10.4 (-2.0 to +22.8) | 0.099 | 6.2 (2.7-13.6) | 3.2 (1.1-7.1) |
| C6 Decision utility | 8/81 | 9.9 (5.1-18.3) | 25/96 | 24.1 (15.3-34.8) | +14.2 (+2.9 to +25.5) | 0.014 | 2.5 (0.7-8.6) | 3.2 (1.1-7.1) |
| C7 Performance with a modality absent | 3/81 | 3.7 (1.3-10.3) | 9/96 | 7.2 (3.8-12.3) | +3.5 (-2.2 to +9.2) | 0.230 | 0.0 (0.0-4.5) | 0.0 (0.0-3.8) |
| C8 Acquisition burden quantified | 1/81 | 1.2 (0.2-6.7) | 2/96 | 2.6 (0.3-8.8) | +1.3 (-2.9 to +5.6) | 0.537 | 0.0 (0.0-4.5) | 0.0 (0.0-3.8) |
| C9 Prospective evaluation | 0/81 | 0.0 (0.0-4.5) | 0/96 | 0.0 (0.0-3.8) | - | - | 0.0 (0.0-4.5) | 0.0 (0.0-3.8) |

Legacy: census of eligible studies (Wilson 95% CI). Update: stratified design (strata I/A/B, FPC N = 386/153/387), full-text unavailable records as within-stratum nonresponse; eligible studies analysed as a domain; Korn-Graubard 95% CI. Difference = Update minus Legacy, design-based contrast (Legacy stratum weight 1, no FPC).

## Extended Data Table 6 | Value claims and claim–evidence mismatch.

| Measure | Legacy, n/N | Legacy, % (95% CI) | Update, n/N (unweighted) | Update, weighted % (95% CI) | Difference, pp (95% CI) | P |
|---|---|---|---|---|---|---|
| Performance only | 13/81 | 16.0 (9.6-25.5) | 13/96 | 12.3 (6.9-20.0) | -3.7 (-13.8 to +6.4) | 0.472 |
| Potential clinical usefulness | 65/81 | 80.2 (70.3-87.5) | 71/96 | 74.0 (62.8-83.3) | -6.2 (-19.3 to +6.8) | 0.350 |
| Claims improved care or outcomes | 3/81 | 3.7 (1.3-10.3) | 12/96 | 13.7 (6.3-24.6) | +9.9 (+0.5 to +19.4) | 0.038 |
| Neither decision utility (C6) nor prospective evaluation (C9) | 73/81 | 90.1 (81.7-94.9) | 71/96 | 75.9 (65.2-84.7) | - | - |
| Claim beyond predictive performance, among studies with neither C6 nor C9 | 61/73 | 83.6 (73.4-90.3) | 60/71 | 85.9 (76.2-92.7) | - | - |
| Claim beyond predictive performance, all studies | 68/81 | 84.0 (74.5-90.4) | 83/96 | 87.7 (80.0-93.1) | - | - |

Legacy: census of eligible studies (Wilson 95% CI). Update: stratified design (strata I/A/B, FPC N = 386/153/387), full-text unavailable records as within-stratum nonresponse; eligible studies analysed as a domain; Korn-Graubard 95% CI.

# Supplementary Information

**Clinical value claims in multimodal AI outpace supporting evidence**

Jiayu Feng

## Contents

## Supplementary Methods

### 1. Search strategy

Database: PubMed. Date searched: 1 October 2026. Publication dates: 28 January 2025 to 30 September 2026, beginning the day after the search date of the earlier review. Records of the Review publication type were excluded in the query. The search returned 926 unique records, which formed the sampling frame of the update corpus (Methods).

```
( "multimodal artificial intelligence"[Title/Abstract] OR "multimodal AI"[Title/Abstract] OR
"integrated data analysis"[Title/Abstract] OR "multi-modal machine learning"[Title/Abstract] OR
"multimodal machine learning"[Title/Abstract] OR "multi-modal deep learning"[Title/Abstract] OR
"multimodal deep learning"[Title/Abstract] OR "fusion data analysis"[Title/Abstract] OR "combined
modalities AI"[Title/Abstract] OR "AI with multiple data sources"[Title/Abstract] OR "multi-input AI"
[Title/Abstract] OR "AI multi-source integration"[Title/Abstract] ) AND ( "diagnostic"[All Fields] OR
"accuracy"[All Fields] OR "effectiveness"[All Fields] OR "performance"[All Fields] OR "sensitivity"
[All Fields] OR "specificity"[All Fields] OR "predictive value"[All Fields] OR "positive predictive
value"[All Fields] OR "negative predictive value"[All Fields] OR "area under the curve"[All Fields] OR
"AUC"[All Fields] OR "receiver operating characteristic"[All Fields] OR "ROC"[All Fields] OR
"precision"[All Fields] OR "assessment"[All Fields] OR "evaluation"[All Fields] ) AND ( "medicine"[All
Fields] OR "healthcare"[All Fields] OR "clinical"[All Fields] OR "patient care"[All Fields] OR
"medical"[All Fields] OR "clinical decision-making"[All Fields] OR "patient diagnosis"[All Fields] OR
"health services"[All Fields] OR "medical treatment"[All Fields] OR "clinical practices"[All Fields] )
NOT Review[Publication Type] AND ("2025/01/28"[Date - Publication] : "2026/09/30"[Date - Publication])
```

### 2. Codebook (version 1.1)

Items were defined before coding of the analysed records began. Every affirmative eligibility or evidence judgment recorded its location in the article (page, table, figure or supplement) and a short verbatim excerpt.

| Item | Question | Values | Coding rule |
|---|---|---|---|
| **Eligibility** | | | |
| Criterion 1, multimodal model | Does the model use two or more independently acquired data sources? | Yes / no | Features derived from the same raw data, sequences within one examination and the same acquisition type at different time points count as one modality. Variables are classified by source. |
| Criterion 2, clinical scope | Does the model use person-level human data to predict or support a clinically meaningful outcome? | Yes / no | Outcomes defined by a validated threshold on a clinical screening instrument are in scope. |
| Criterion 3, qualifying comparator | Is a model built on a strict subset of the modalities reported on the same data and outcome? | Yes / no | A separately fitted unimodal model, a modality-withheld ablation or a nested addition. Image-derived measurements belong to imaging, so a nominally clinical model containing them is not unimodal. |
| Location of the comparison | Where is the qualifying comparison first reported? | Abstract / main text / table / figure / supplement | Abstract if it appears anywhere in the abstract; otherwise the first full-text location. |
| **Evidence items** | | | |
| C1, gain quantifiable | Are multimodal and comparator performance reported for the same outcome, metric and evaluation set? | Yes / no | Met when yes. |

| Item | Question | Values | Coding rule |
|---|---|---|---|
| C2, complete unimodal benchmarking | Does every constituent modality have its own unimodal model? | Yes / no / unclear | Met when every constituent modality has its own separately fitted unimodal model on the same evaluation set. Masking inputs of the trained multimodal model does not count. Nested-addition designs will usually not meet the item. |
| C3, uncertainty of the gain | Is uncertainty reported for the difference itself? | Yes / no | Met by a confidence interval of the difference, a paired test or a bootstrap of the difference. Separate intervals for each model do not count. |
| C4, external validation of the gain | On what data is the gain validated? | None / temporal / external | Met only when external: both models evaluated on data from a structurally distinct site, population or setting, including a separate-source public dataset. A later period at the same site is temporal. Validating the multimodal model alone does not count. |
| C5, calibration | Is calibration of the multimodal model assessed? | Yes / no | Met by a calibration plot, calibration intercept or slope, or an observed-versus-predicted assessment. A Brier score alone does not count. |
| C6, decision utility | Is the decision utility of the model reported? | Yes / no | Met by decision-curve or net-benefit analysis, or an explicit threshold-based utility or cost-benefit analysis. Net reclassification or integrated discrimination improvement alone, or sensitivity and specificity at a threshold alone, do not count. |
| C7, performance with a modality absent | Is the trained multimodal model evaluated with an entire modality absent? | Yes / no | Met when performance is reported with an entire modality absent (masking, dropout or a missing-modality evaluation) in the already-trained multimodal model. Switching to a separately trained model, or item-level imputation within a modality, does not count. |
| C8, acquisition burden | Is the incremental burden or cost of acquiring the added modality addressed? | None / discussed / quantified | Met only when quantified, and only for the added modality. Computational cost or inference time does not count. |
| C9, prospective evaluation | Is the model evaluated prospectively? | None / prospective observational / silent deployment / interventional trial | Met by any value other than none. Requires predictions generated prospectively; prospective data collection alone, or retrospective real-world data, does not count. |
| **Primary comparison** | | | |
| Primary metric | Which metric is primary? | AUROC / C-index / AUPRC / accuracy / F1 / sensitivity-specificity / other | The authors' designated metric; otherwise the metric of the main between-model comparison reported in the abstract; otherwise AUROC. |
| Model values | Values of the primary multimodal model and the strongest eligible unimodal comparator on the primary metric | Number | Same outcome, metric and evaluation set for both, with page and excerpt. |
| Test-set size and events | Number of patients contributing predictions, and number of events | Number / not reported | For cross-validation, the number of patients contributing out-of-fold predictions. |
| Basis of the primary model | How was the primary multimodal model identified? | Author-designated / four-step hierarchy | Never selected by its score; the hierarchy always resolves (Methods). |
| AUROC of both models | AUROC of the primary multimodal model and the strongest eligible comparator, if reported | Number / not reported | Used for ΔAUROC. |
| **Value claim** | | | |
| Highest explicit value claim | What is the strongest claim of value in the abstract or conclusion? | Performance only / potential clinical usefulness / improved care or outcomes | Hedged statements (may, could, potential) count as potential clinical usefulness. An explicit claim of improved care or outcomes is the highest level, and a caveat clause does not downgrade it. The item records |

| Item | Question | Values | Coding rule |
|---|---|---|---|
| | | | the height of the claim, not whether the evidence supports it. |
| **Covariates** | | | |
| Modality types | Modalities used by the primary multimodal model | Imaging / physiological signal / structured clinical or EHR / free text / omics / audio or video / other | One indicator per type. |
| Task | Clinical task | Diagnosis or detection / prognosis or risk / treatment response / other clinical | Monitoring or state-classification tasks are other clinical. |
| Specialty | Specialty of the primary target condition | Oncology / cardiovascular / neurology-psychiatry / ophthalmology / respiratory-critical care / other | |
| Data source | Source of the data | Public / institutional / both | |
| Model class | Class of the primary model | Classical machine learning / deep learning / foundation model or LLM | |
| Data-collection design | Design of data collection | Retrospective / prospective / mixed / not reported | |
| Flags | Preprint; methods or resource paper; borderline multimodal status | Yes / no | Used for sensitivity analyses. |

## 3. Operational clarifications

The written decisions below were applied to all records. None changed the evidence items or the primary-comparison rule.

1. **Figure-derived values.** When the values of the primary comparison were reported only in a figure, the best discernible value was read from the figure, marked as approximate and figure-derived, and recorded with no more precision than the figure allowed.
2. **Multiple outcomes or tasks.** The author-designated primary outcome or task was used; failing that, the first listed in the main results. The primary-comparison rule was then applied within that outcome or task. Values were not averaged across tasks, cancers or outcomes unless the authors reported only an aggregate.
3. **Provenance of image-derived variables.** A variable was assigned to imaging or pathology only when the full text stated that it was derived from that modality. Variable names or clinical knowledge alone (for example, tumour stage or size) were not sufficient. The same threshold applied to both corpora.
4. **Methods or resource papers.** A paper was flagged as a methods or resource paper when it positioned its main contribution as a generalizable method, software tool, framework, dataset or resource, benchmark or general model evaluation. A paper modelling one clinical population, disease or care setting was not flagged. Whether the data were public was irrelevant.
5. **Data-collection design.** The authors' self-reported design took precedence. Without one, participants recruited for the reported study counted as prospective, secondary use of existing data as retrospective, both as mixed, and otherwise as not reported.
6. **Modality types.** Only the modalities used by the primary multimodal model were recorded, not every modality analysed in the study.
7. **Test-set size.** When the test-set size was not reported but could be computed from the reported split ratio and total sample, the computed value was recorded as approximate; otherwise it was recorded as not reported.
8. **Specialty.** Specialty was classified by the study's primary target condition, not by the authors' department: cancers (including skin cancer) as oncology; stroke, neurodegenerative and psychiatric conditions as neurology-psychiatry; COVID-19, respiratory conditions and critical care as respiratory-critical care; cardiovascular conditions as cardiovascular; eye conditions as ophthalmology; otherwise other.

9. **Data source.** Public data were openly downloadable datasets, challenge or benchmark datasets, or third-party research resources with a standard external access pathway, even when access was controlled (for example, TCGA, UK Biobank, ADNI or MIMIC). Institutional data came from the authors' institution, network or a closed consortium, including data available only on request. Both applied when the primary model used both kinds of source.

### 4. Coverage audit of a second database and sensitivity analysis

To estimate how many eligible studies a second database would add, and whether they would change the estimates, we audited Embase. The audit protocol was fixed before its sample was drawn, and the sensitivity analysis before any eligible study was coded.

**Search.** Embase was searched on 7 October 2026 with the concepts of the PubMed search translated to Embase syntax, for the same publication dates (28 January 2025 to 30 September 2026). The query excluded reviews and records from the ClinicalTrials.gov collection:

```
(
  ('multimodal artificial intelligence':ti,ab OR 'multimodal AI':ti,ab OR 'integrated data
analysis':ti,ab
   OR 'multi-modal machine learning':ti,ab OR 'multimodal machine learning':ti,ab OR 'multi-modal deep
learning':ti,ab
   OR 'multimodal deep learning':ti,ab OR 'fusion data analysis':ti,ab OR 'combined modalities
AI':ti,ab
   OR 'AI with multiple data sources':ti,ab OR 'multi-input AI':ti,ab OR 'AI multi-source
integration':ti,ab)
  AND (diagnostic OR accuracy OR effectiveness OR performance OR sensitivity OR specificity OR
'predictive value'
   OR 'positive predictive value' OR 'negative predictive value' OR 'area under the curve' OR AUC
   OR 'receiver operating characteristic' OR ROC OR precision OR assessment OR evaluation)
  AND (medicine OR healthcare OR clinical OR 'patient care' OR medical OR 'clinical decision-making'
   OR 'patient diagnosis' OR 'health services' OR 'medical treatment' OR 'clinical practices')
)
NOT 'review':it
AND [2025-01-28 TO 2026-09-30]/pd
NOT 'clinical trial':dtype
```

**Records.** The search returned 1,219 records, 1,182 after removing 37 duplicates within Embase. Records were matched to the 926-record PubMed frame by PMID, DOI and normalized title, and titles with a similarity of 0.90 or more were reviewed by hand. Of the 1,182 records, 584 matched the PubMed frame. Of the other 598, nine were preprints or conference abstracts of studies in the frame, which left 589 records not linked to any study in it. The audit population was the 352 of these that were full reports (article, article in press or preprint), since the main study required a full report. Conference abstracts, letters, editorials, notes, errata and short surveys were excluded by type. The audit measures what a second database adds to the PubMed search frame, not the coverage of either database.

**Sample, eligibility and coding.** We drew a simple random sample of 100 of the 352 records with a fixed seed and assessed every one; none was replaced. The eligibility criteria were those of the main study (Methods). At title and abstract, a record was excluded only when it clearly failed the multimodal or clinical-scope criterion or was a review, protocol, perspective or editorial. A missing comparator never excluded a record at this stage. A record counted as an additional study only when the same study was not already in the corpus, legacy studies included. Full texts that could not be obtained were kept as nonrespondents. Eligible additional studies were coded with codebook version 1.1, the primary-comparison rule and the clarifications above, as in the main study.

**Estimation.** The number of eligible additional studies among the 352 records was estimated as $352z/n$ with an exact 95% interval from the hypergeometric distribution, and expressed relative to the 434 eligible studies estimated for the PubMed frame. For the sensitivity analysis, the 352 records formed a fourth update stratum (stratum E), disjoint from the PubMed frame. Respondents were the sampled records with full text, the eligible studies formed the domain, and the design weight was 352 divided by the number of respondents, with a finite-population correction. The legacy corpus and strata I, A and B were unchanged. Four prespecified sets of estimates were recomputed with the estimators of the primary analysis: prevalence of the evidence items, in the update period and pooled, with legacy–update differences; the share of studies in which the multimodal model beat its strongest comparator; value claims; and the primary regression of ΔAUROC. Differences between the two analyses are described, not tested.

**Results.** Of the 100 sampled records, 62 were excluded at title and abstract and full text was unavailable for 3. Of the 35 full texts assessed, 20 were eligible additional studies, 12 of them preprints. This corresponds to an estimated 73 eligible additional studies among the 352 records (95% CI, 50–100), 16.7% of the number estimated for the PubMed frame. Counting the unavailable records as eligible gives 81 (58–109) (Supplementary Table S11). With stratum E added, the update-period prevalence of each evidence item changed by 4.5 percentage points or less, the largest change being for uncertainty of the gain (41.2% to 36.7%). The share of update studies in which the multimodal model beat its strongest comparator went from 89.3% to 87.9% (Supplementary Tables S12 and S13 and Supplementary Fig. 1). The coefficient for complete benchmarking was −0.041 (95% CI, −0.065 to −0.017), against −0.030 in the primary analysis (Supplementary Table S14). Two results lost statistical significance: the legacy–update increase in claims of improved care (+8.0 percentage points; 95% CI, −0.4 to 16.4) and the coefficient for log test-set size (P = 0.060).

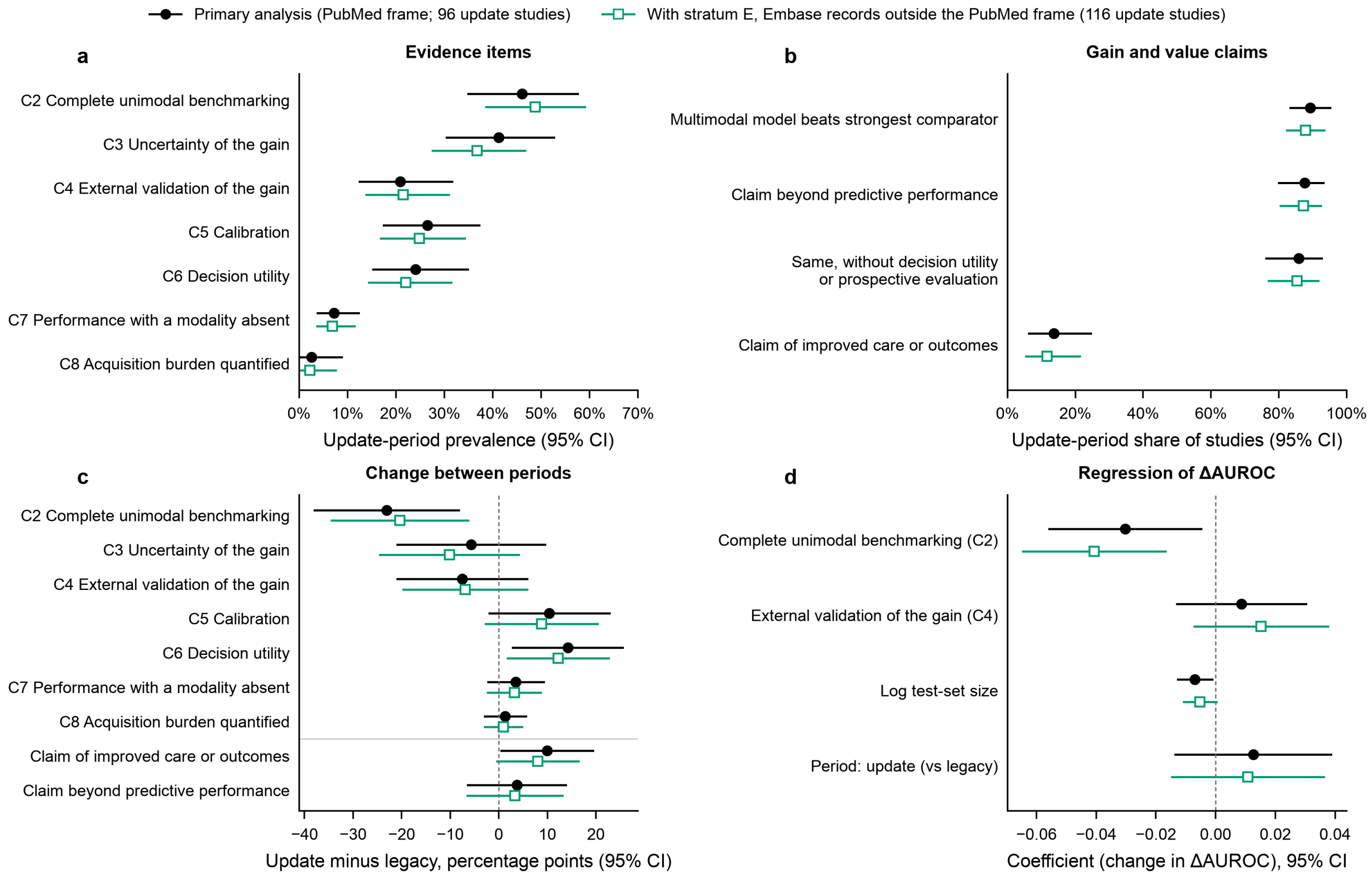

**Supplementary Fig. 1 | Estimates with and without stratum E.** Primary analysis (PubMed frame, strata I, A and B) and the analysis adding stratum E (352 Embase full reports outside the PubMed frame, 20 eligible studies). **a**, Update-period prevalence of evidence items C2–C8; C1 was met by every study and C9 by none in both analyses. **b**, Share of update studies in which the multimodal model beat the strongest comparator on the primary metric, and value claims. **c**, Update minus legacy, design-based contrasts. **d**, Survey-weighted regression of ΔAUROC (110 studies in the primary analysis, 125 with stratum E). Korn–Graubard 95% CIs for prevalences; Wald 95% CIs for the outperformance share, contrasts and coefficients.

## Supplementary Tables

**Supplementary Table S1 |** Mean ΔAUROC by study feature.

| Feature | Studies with / without (n) | Mean ΔAUROC, with | Mean ΔAUROC, without | Difference (95% CI) | P |
|---|---|---|---|---|---|
| Complete unimodal benchmarking (C2) | 79 / 48 | 0.046 | 0.080 | −0.034 (−0.057 to −0.011) | 0.004 |
| Uncertainty of the gain (C3) | 64 / 63 | 0.057 | 0.063 | −0.006 (−0.029 to 0.017) | 0.600 |
| External validation of the gain (C4) | 35 / 92 | 0.063 | 0.060 | 0.003 (−0.021 to 0.026) | 0.831 |
| Temporal or external validation | 44 / 83 | 0.063 | 0.059 | 0.004 (−0.018 to 0.026) | 0.728 |
| Calibration (C5) | 37 / 90 | 0.067 | 0.057 | 0.010 (−0.014 to 0.033) | 0.425 |
| Decision utility (C6) | 31 / 96 | 0.072 | 0.057 | 0.015 (−0.008 to 0.038) | 0.198 |
| Performance with a modality absent (C7) | 6 / 121 | 0.046 | 0.061 | −0.016 (−0.052 to 0.021) | 0.394 |
| Comparison reported in abstract | 101 / 26 | 0.057 | 0.072 | −0.015 (−0.053 to 0.022) | 0.427 |
| Primary model designated by authors | 99 / 28 | 0.065 | 0.043 | 0.022 (−0.007 to 0.052) | 0.131 |
| Cross-validated evaluation | 30 / 97 | 0.045 | 0.065 | −0.019 (−0.042 to 0.004) | 0.103 |
| Public data | 32 / 95 | 0.053 | 0.063 | −0.011 (−0.035 to 0.014) | 0.402 |
| Methods or resource paper | 7 / 120 | 0.026 | 0.063 | −0.037 (−0.052 to −0.023) | <0.001 |
| Deep learning or foundation model | 103 / 24 | 0.059 | 0.066 | −0.007 (−0.036 to 0.022) | 0.651 |
| Prospective or mixed data | 14 / 113 | 0.042 | 0.063 | −0.021 (−0.051 to 0.009) | 0.161 |
| Claim of improved care | 11 / 116 | 0.080 | 0.059 | 0.021 (−0.003 to 0.045) | 0.081 |

127 studies reporting AUROC for both models. Means are pooled and design-weighted; the difference (with minus without) is from a survey-weighted linear regression on a single indicator, without adjustment. Counts are unweighted. C2 and C4 are terms of the primary regression; the other features are exploratory.

**Supplementary Table S2 |** Post hoc alternative specifications of the primary regression.

| Specification | Term | Coefficient (95% CI) | P | n |
|---|---|---|---|---|
| Unweighted least squares | Intercept | 0.1210 (0.0713 to 0.1707) | <0.001 | 110 |
| Unweighted least squares | External validation of the gain (C4) | 0.0032 (−0.0239 to 0.0302) | 0.817 | 110 |
| Unweighted least squares | Complete unimodal benchmarking (C2) | −0.0266 (−0.0520 to −0.0012) | 0.040 | 110 |
| Unweighted least squares | Log test-set size | −0.0099 (−0.0180 to −0.0018) | 0.018 | 110 |
| Unweighted least squares | Period: update (vs legacy) | 0.0092 (−0.0157 to 0.0341) | 0.464 | 110 |
| Excluding influential studies (Cook's D > 4/n) | Intercept | 0.0864 (0.0540 to 0.1188) | <0.001 | 106 |
| Excluding influential studies (Cook's D > 4/n) | External validation of the gain (C4) | 0.0060 (−0.0098 to 0.0217) | 0.454 | 106 |
| Excluding influential studies (Cook's D > 4/n) | Complete unimodal benchmarking (C2) | −0.0240 (−0.0412 to −0.0068) | 0.007 | 106 |
| Excluding influential studies (Cook's D > 4/n) | Log test-set size | −0.0064 (−0.0111 to −0.0017) | 0.008 | 106 |
| Excluding influential studies (Cook's D > 4/n) | Period: update (vs legacy) | 0.0236 (0.0109 to 0.0363) | <0.001 | 106 |
| Excluding the 18-patient study | Intercept | 0.0838 (0.0487 to 0.1189) | <0.001 | 109 |
| Excluding the 18-patient study | External validation of the gain (C4) | 0.0115 (−0.0085 to 0.0314) | 0.258 | 109 |
| Excluding the 18-patient study | Complete unimodal benchmarking (C2) | −0.0263 (−0.0447 to −0.0079) | 0.005 | 109 |
| Excluding the 18-patient study | Log test-set size | −0.0054 (−0.0104 to −0.0003) | 0.038 | 109 |
| Excluding the 18-patient study | Period: update (vs legacy) | 0.0233 (0.0080 to 0.0386) | 0.003 | 109 |
| ΔAUROC winsorized at the 5th and 95th percentiles | Intercept | 0.0856 (0.0535 to 0.1177) | <0.001 | 110 |
| ΔAUROC winsorized at the 5th and 95th percentiles | External validation of the gain (C4) | 0.0081 (−0.0087 to 0.0248) | 0.342 | 110 |
| ΔAUROC winsorized at the 5th and 95th percentiles | Complete unimodal benchmarking (C2) | −0.0245 (−0.0413 to −0.0078) | 0.005 | 110 |
| ΔAUROC winsorized at the 5th and 95th percentiles | Log test-set size | −0.0055 (−0.0102 to −0.0008) | 0.023 | 110 |
| ΔAUROC winsorized at the 5th and 95th percentiles | Period: update (vs legacy) | 0.0208 (0.0070 to 0.0347) | 0.004 | 110 |
| Outcome: relative gain, ΔAUROC/(1 − comparator AUROC) | Intercept | 0.3915 (0.2476 to 0.5354) | <0.001 | 110 |
| Outcome: relative gain, ΔAUROC/(1 − comparator AUROC) | External validation of the gain (C4) | 0.0191 (−0.0622 to 0.1005) | 0.642 | 110 |
| Outcome: relative gain, ΔAUROC/(1 − comparator AUROC) | Complete unimodal benchmarking (C2) | −0.0691 (−0.1464 to 0.0083) | 0.079 | 110 |
| Outcome: relative gain, ΔAUROC/(1 − comparator AUROC) | Log test-set size | −0.0210 (−0.0434 to 0.0014) | 0.066 | 110 |
| Outcome: relative gain, ΔAUROC/(1 − comparator AUROC) | Period: update (vs legacy) | 0.0828 (0.0068 to 0.1587) | 0.033 | 110 |

| Specification | Term | Coefficient (95% CI) | P | n |
|---|---|---|---|---|
| Median regression | Intercept | 0.0780 (0.0062 to 0.1497) | 0.036 | 110 |
| Median regression | External validation of the gain (C4) | 0.0073 (−0.0291 to 0.0437) | 0.694 | 110 |
| Median regression | Complete unimodal benchmarking (C2) | −0.0257 (−0.0770 to 0.0256) | 0.328 | 110 |
| Median regression | Log test-set size | −0.0044 (−0.0123 to 0.0035) | 0.276 | 110 |
| Median regression | Period: update (vs legacy) | 0.0115 (−0.0119 to 0.0348) | 0.339 | 110 |
| C4 counting temporal validation | Intercept | 0.1043 (0.0510 to 0.1577) | <0.001 | 110 |
| C4 counting temporal validation | External validation of the gain (C4) | 0.0128 (−0.0057 to 0.0314) | 0.174 | 110 |
| C4 counting temporal validation | Complete unimodal benchmarking (C2) | −0.0306 (−0.0503 to −0.0109) | 0.003 | 110 |
| C4 counting temporal validation | Log test-set size | −0.0071 (−0.0126 to −0.0016) | 0.012 | 110 |
| C4 counting temporal validation | Period: update (vs legacy) | 0.0133 (−0.0123 to 0.0389) | 0.304 | 110 |
| Adding uncertainty reporting (C3) | Intercept | 0.1072 (0.0518 to 0.1626) | <0.001 | 110 |
| Adding uncertainty reporting (C3) | External validation of the gain (C4) | 0.0105 (−0.0111 to 0.0321) | 0.338 | 110 |
| Adding uncertainty reporting (C3) | Complete unimodal benchmarking (C2) | −0.0299 (−0.0499 to −0.0100) | 0.004 | 110 |
| Adding uncertainty reporting (C3) | Uncertainty of the gain (C3) | −0.0050 (−0.0238 to 0.0139) | 0.604 | 110 |
| Adding uncertainty reporting (C3) | Log test-set size | −0.0069 (−0.0126 to −0.0012) | 0.018 | 110 |
| Adding uncertainty reporting (C3) | Period: update (vs legacy) | 0.0125 (−0.0133 to 0.0383) | 0.339 | 110 |
| Outcome: difference in logit AUROC | Intercept | 0.7467 (0.4517 to 1.0417) | <0.001 | 110 |
| Outcome: difference in logit AUROC | External validation of the gain (C4) | 0.0417 (−0.1281 to 0.2115) | 0.627 | 110 |
| Outcome: difference in logit AUROC | Complete unimodal benchmarking (C2) | −0.1675 (−0.3216 to −0.0135) | 0.033 | 110 |
| Outcome: difference in logit AUROC | Log test-set size | −0.0460 (−0.0862 to −0.0058) | 0.025 | 110 |
| Outcome: difference in logit AUROC | Period: update (vs legacy) | 0.1204 (−0.0336 to 0.2744) | 0.124 | 110 |
| Adjusted for number of modality types, modality types and task | Intercept | 0.0719 (−0.0024 to 0.1461) | 0.058 | 110 |
| Adjusted for number of modality types, modality types and task | External validation of the gain (C4) | 0.0011 (−0.0196 to 0.0218) | 0.914 | 110 |
| Adjusted for number of modality types, modality types and task | Complete unimodal benchmarking (C2) | −0.0261 (−0.0496 to −0.0026) | 0.030 | 110 |
| Adjusted for number of modality types, modality types and task | Log test-set size | −0.0063 (−0.0132 to 0.0005) | 0.070 | 110 |

| Specification | Term | Coefficient (95% CI) | P | n |
|---|---|---|---|---|
| Adjusted for number of modality types, modality types and task | Period: update (vs legacy) | 0.0132 (−0.0134 to 0.0399) | 0.327 | 110 |
| Adjusted for number of modality types, modality types and task | Two or more modality types (vs one) | −0.0502 (−0.0973 to −0.0032) | 0.036 | 110 |
| Adjusted for number of modality types, modality types and task | Imaging | 0.0148 (−0.0127 to 0.0422) | 0.288 | 110 |
| Adjusted for number of modality types, modality types and task | Structured clinical or EHR data | 0.0539 (0.0109 to 0.0970) | 0.015 | 110 |
| Adjusted for number of modality types, modality types and task | Omics | 0.0200 (−0.0169 to 0.0569) | 0.285 | 110 |
| Adjusted for number of modality types, modality types and task | Free text | 0.0079 (−0.0381 to 0.0539) | 0.734 | 110 |
| Adjusted for number of modality types, modality types and task | Task: prognosis or risk (vs diagnosis) | 0.0215 (−0.0063 to 0.0493) | 0.127 | 110 |
| Adjusted for number of modality types, modality types and task | Task: treatment response or other (vs diagnosis) | 0.0401 (0.0120 to 0.0682) | 0.006 | 110 |
| Without log test-set size | Intercept | 0.0750 (0.0431 to 0.1068) | <0.001 | 127 |
| Without log test-set size | External validation of the gain (C4) | 0.0082 (−0.0131 to 0.0296) | 0.446 | 127 |
| Without log test-set size | Complete unimodal benchmarking (C2) | −0.0344 (−0.0572 to −0.0116) | 0.003 | 127 |
| Without log test-set size | Period: update (vs legacy) | 0.0039 (−0.0213 to 0.0291) | 0.759 | 127 |
| Outcome: gain on the study's primary metric | Intercept | 0.0978 (0.0529 to 0.1427) | <0.001 | 145 |
| Outcome: gain on the study's primary metric | External validation of the gain (C4) | 0.0034 (−0.0144 to 0.0211) | 0.707 | 145 |
| Outcome: gain on the study's primary metric | Complete unimodal benchmarking (C2) | −0.0230 (−0.0422 to −0.0038) | 0.019 | 145 |
| Outcome: gain on the study's primary metric | Log test-set size | −0.0056 (−0.0100 to −0.0012) | 0.012 | 145 |
| Outcome: gain on the study's primary metric | Period: update (vs legacy) | 0.0056 (−0.0165 to 0.0277) | 0.616 | 145 |
| Outcome: gain on the study's primary metric, without log test-set size | Intercept | 0.0742 (0.0477 to 0.1008) | <0.001 | 169 |
| Outcome: gain on the study's primary metric, without log test-set size | External validation of the gain (C4) | 0.0036 (−0.0131 to 0.0202) | 0.674 | 169 |
| Outcome: gain on the study's primary metric, without log test-set size | Complete unimodal benchmarking (C2) | −0.0295 (−0.0473 to −0.0117) | 0.001 | 169 |
| Outcome: gain on the study's primary metric, without log test-set size | Period: update (vs legacy) | 0.0002 (−0.0213 to 0.0218) | 0.983 | 169 |

Outcome ΔAUROC regressed on complete unimodal benchmarking (C2), external validation of the gain (C4), log test-set size and period unless stated. Survey-weighted specifications use the primary design. Unweighted least squares ignores the design. Median regression uses the design weights with bootstrap standard errors (2,000 resamples); its 95% CI is the estimate ± 1.96 s.e. The adjusted specification adds the number of modality types (two or more versus one), indicators for imaging, structured clinical or EHR data, omics and free text, and the prediction task. The logit outcome is logit(multimodal AUROC) − logit(comparator AUROC). The primary-metric outcome is the multimodal minus the comparator value on each study's own primary metric, all of which lie between 0 and 1; it is available for 169 studies. The relative-gain, logit and primary-metric outcomes are on different scales from ΔAUROC. P values are descriptive.

**Supplementary Table S3 |** Bootstrap of the primary regression.

| Term | Bootstrap median | 2.5th to 97.5th percentile | Resamples below zero, % |
|---|---|---|---|
| Intercept | 0.1025 | 0.0448 to 0.1663 | 0.0 |
| External validation of the gain (C4) | 0.0088 | −0.0149 to 0.0341 | 24.7 |
| Complete unimodal benchmarking (C2) | −0.0297 | −0.0572 to −0.0023 | 98.4 |
| Log test-set size | −0.0066 | −0.0132 to 0.0006 | 96.5 |
| Period: update (vs legacy) | 0.0128 | −0.0197 to 0.0362 | 18.1 |

2,000 resamples of the 110 studies with replacement; in each resample the primary model was refitted by weighted least squares with the design weights. Post hoc.

**Supplementary Table S4 |** ΔAUROC by comparator AUROC.

| Comparator AUROC | Studies (n) | Median ΔAUROC | Median relative gain |
|---|---|---|---|
| 0.70 or lower | 16 | 0.127 | 0.32 |
| Above 0.70 to 0.80 | 45 | 0.046 | 0.20 |
| Above 0.80 to 0.90 | 48 | 0.042 | 0.28 |
| Above 0.90 | 18 | 0.023 | 0.38 |

127 studies reporting AUROC for both models. Relative gain is ΔAUROC/(1 − comparator AUROC). Medians are unweighted. Exploratory.

**Supplementary Table S5 |** Complete benchmarking and decision utility by screening stratum, update period.

| Stratum | Eligible studies (n) | Design weight | Complete unimodal benchmarking (C2), % | Decision utility (C6), % |
|---|---|---|---|---|
| I, included at screening | 83 | 3.48 | 55.4 | 27.7 |
| A, included by the first binary pass, excluded at adjudication | 8 | 7.65 | 37.5 | 12.5 |
| B, excluded by both binary passes | 5 | 16.83 | 20.0 | 20.0 |

Unweighted proportions within each stratum of the 96 eligible update studies. Design weights are $N_h/n_h$ among respondents (Methods).

**Supplementary Table S6 |** Complete unimodal benchmarking by number of data sources.

| Corpus | Studies (n) | Three or more data sources, % | C2 with two data sources, % | C2 with three or more, % | C2 overall, % |
|---|---|---|---|---|---|
| Legacy | 81 | 12.3 | 65.2 | 80.0 | 69.1 |
| Update | 96 | 11.5 | 47.9 | 36.4 | 52.1 |

Unweighted proportions of eligible studies. Number of data sources refers to the modality types used by the primary multimodal model.

**Supplementary Table S7 |** Update-period prevalence of evidence items under alternative weighting.

| Evidence item | Design weights (primary), % | No weights, % | Design weights excluding stratum B, % | Stratum I only, % |
|---|---|---|---|---|
| Complete unimodal benchmarking (C2) | 46.0 | 52.1 | 52.3 | 55.4 |
| Uncertainty of the gain (C3) | 41.2 | 45.8 | 46.3 | 48.2 |
| External validation of the gain (C4) | 20.9 | 21.9 | 21.1 | 22.9 |
| Calibration (C5) | 26.5 | 29.2 | 28.0 | 31.3 |
| Decision utility (C6) | 24.1 | 26.0 | 25.0 | 27.7 |
| Performance with a modality absent (C7) | 7.2 | 9.4 | 8.9 | 10.8 |
| Acquisition burden quantified (C8) | 2.6 | 2.1 | 3.2 | 1.2 |

96 eligible update studies. C1 was met by every study and C9 by none under every weighting. See also Extended Data Fig. 3.

**Supplementary Table S8 |** Co-occurrence of evidence items.

| Studies meeting | C2 | C3 | C4 | C5 | C6 | C7 | C8 |
|---|---|---|---|---|---|---|---|
| Complete unimodal benchmarking (C2) | – | 47 | 24 | 18 | 16 | 4 | 3 |
| Uncertainty of the gain (C3) | 55 | – | 29 | 32 | 29 | 10 | 0 |
| External validation of the gain (C4) | 53 | 55 | – | 29 | 42 | 16 | 1 |
| Calibration (C5) | 35 | 54 | 26 | – | 58 | 6 | 0 |
| Decision utility (C6) | 36 | 56 | 42 | 66 | – | 0 | 0 |
| Performance with a modality absent (C7) | 26 | 64 | 54 | 23 | 0 | – | 0 |
| Acquisition burden quantified (C8) | 71 | 8 | 8 | 0 | 0 | 0 | – |

Share (%) of studies meeting the row item that also meet the column item. Pooled across corpora and design-weighted (177 studies). Exploratory.

**Supplementary Table S9 |** Evidence profiles from hierarchical clustering.

| Cluster | Studies (n) | Legacy (n) | C2, % | C3, % | C4, % | C5, % | C6, % | C7, % | C8, % | Mean ΔAUROC (n) | Claim beyond performance, % |
|---|---|---|---|---|---|---|---|---|---|---|---|
| 1 | 127 | 65 | 70 | 36 | 15 | 2 | 2 | 6 | 2 | 0.053 (84) | 84 |
| 2 | 24 | 5 | 0 | 42 | 21 | 83 | 79 | 0 | 0 | 0.073 (20) | 96 |
| 3 | 17 | 6 | 94 | 100 | 65 | 88 | 71 | 0 | 0 | 0.049 (17) | 88 |
| 4 | 8 | 4 | 0 | 100 | 100 | 38 | 0 | 62 | 0 | 0.070 (6) | 62 |
| 5 | 1 | 1 | 100 | 100 | 100 | 0 | 0 | 0 | 100 | – (0) | 100 |

Average-linkage hierarchical clustering of the 177 studies on items C2–C8 (Gower distance on binary indicators), cut at five clusters; average silhouette width 0.38. Shares and means are unweighted. Exploratory.

**Supplementary Table S10 |** Value claims by evidence item.

| Evidence item | Status | Studies (n) | Claim beyond predictive performance, % | Claim of improved care or outcomes, % |
|---|---|---|---|---|
| External validation of the gain (C4) | Not met | 133 | 85.9 | 8.7 |
| External validation of the gain (C4) | Met | 44 | 91.2 | 24.0 |
| Calibration (C5) | Not met | 136 | 85.1 | 12.0 |
| Calibration (C5) | Met | 41 | 93.0 | 12.4 |
| Decision utility (C6) | Not met | 144 | 85.4 | 6.5 |
| Decision utility (C6) | Met | 33 | 92.9 | 32.2 |

Pooled across corpora and design-weighted; counts are unweighted. No study in either group evaluated its model prospectively. Exploratory.

**Supplementary Table S11 |** Eligible additional studies among the 352 Embase full reports outside the PubMed frame.

| Estimate | Eligible / denominator | Estimated eligible among 352 (exact 95% CI) | Relative to the 434 eligible studies estimated for the PubMed frame, % (95% CI) |
|---|---|---|---|
| Primary (assessed records) | 20 / 97 | 73 (50–100) | 16.7 (11.5–23.0) |
| Bound: unavailable records ineligible | 20 / 100 | 70 (49–97) | 16.2 (11.3–22.4) |
| Bound: unavailable records eligible | 23 / 100 | 81 (58–109) | 18.6 (13.4–25.1) |

Simple random sample of 100 of 352 records: 62 excluded at title and abstract, 38 assessed on full text, of which 3 were unavailable. Exact intervals from the hypergeometric distribution (finite population of 352).

**Supplementary Table S12 |** Evidence items with and without stratum E.

| Measure | Primary, update n/N | Primary, update % | With stratum E, update n/N | With stratum E, update % | Primary, pooled % | With stratum E, pooled % | Primary, update − legacy, pp | With stratum E, update − legacy, pp |
|---|---|---|---|---|---|---|---|---|
| Gain quantifiable (C1) | 96/96 | 100.0 (96.2–100.0) | 116/116 | 100.0 (96.9–100.0) | 100.0 (97.9–100.0) | 100.0 (98.1–100.0) | – | – |
| Complete unimodal benchmarking (C2) | 50/96 | 46.0 (34.9–57.5) | 63/116 | 48.7 (38.6–59.0) | 49.7 (39.8–59.5) | 51.6 (42.5–60.5) | −23.1 (−38.0 to −8.3) | −20.4 (−34.5 to −6.3) |
| Uncertainty of the gain (C3) | 44/96 | 41.2 (30.5–52.6) | 46/116 | 36.7 (27.6–46.7) | 42.1 (32.8–51.8) | 38.1 (30.0–46.8) | −5.7 (−20.9 to +9.5) | −10.2 (−24.4 to +4.1) |
| External validation of the gain (C4) | 21/96 | 20.9 (12.5–31.6) | 26/116 | 21.5 (13.8–30.8) | 22.0 (14.7–31.0) | 22.4 (15.6–30.5) | −7.5 (−20.9 to +5.8) | −6.9 (−19.7 to +5.8) |
| Calibration (C5) | 28/96 | 26.5 (17.5–37.2) | 31/116 | 24.8 (16.9–34.2) | 24.8 (17.1–33.9) | 23.6 (16.7–31.8) | +10.4 (−2.0 to +22.8) | +8.8 (−2.8 to +20.3) |
| Decision utility (C6) | 25/96 | 24.1 (15.3–34.8) | 27/116 | 22.1 (14.4–31.4) | 21.8 (14.4–30.9) | 20.4 (13.7–28.5) | +14.2 (+2.9 to +25.5) | +12.2 (+1.8 to +22.6) |
| Performance with a modality absent (C7) | 9/96 | 7.2 (3.8–12.3) | 10/116 | 6.9 (3.7–11.4) | 6.7 (3.7–10.9) | 6.5 (3.7–10.4) | +3.5 (−2.2 to +9.2) | +3.2 (−2.3 to +8.7) |
| Acquisition burden quantified (C8) | 2/96 | 2.6 (0.3–8.8) | 2/116 | 2.2 (0.3–7.6) | 2.4 (0.4–7.5) | 2.1 (0.3–6.6) | +1.3 (−2.9 to +5.6) | +1.0 (−2.9 to +4.8) |
| Prospective evaluation (C9) | 0/96 | 0.0 (0.0–3.8) | 0/116 | 0.0 (0.0–3.1) | 0.0 (0.0–2.1) | 0.0 (0.0–1.9) | – | – |

Primary: strata I, A and B of the PubMed frame. With stratum E: the 352 Embase full reports outside the PubMed frame added as a fourth update stratum (20 eligible studies; weight 352/97 with a finite-population correction). Counts are unweighted; percentages are design-weighted with 95% CIs (Korn–Graubard for prevalences, Wald for the outperformance share). Pooled = legacy census (weight 1) plus the update design. Differences are design-based contrasts, not shown for items met by all or no studies. Studies without a numeric value for both models count as not showing a gain.

**Supplementary Table S13 |** Outperformance and value claims with and without stratum E.

| Measure | Primary, update n/N | Primary, update % | With stratum E, update n/N | With stratum E, update % | Primary, pooled % | With stratum E, pooled % | Primary, update − legacy, pp | With stratum E, update − legacy, pp |
|---|---|---|---|---|---|---|---|---|
| Multimodal model beats the strongest comparator | 85/96 | 89.3 (83.4–95.1) | 101/116 | 87.9 (82.4–93.4) | 88.6 (83.5–93.7) | 87.6 (82.7–92.4) | – | – |
| Claim: performance only | 13/96 | 12.3 (6.9–20.0) | 16/116 | 12.7 (7.6–19.5) | 12.9 (8.0–19.3) | 13.2 (8.6–19.0) | −3.7 (−13.8 to +6.4) | −3.3 (−13.1 to +6.5) |
| Claim: potential clinical usefulness | 71/96 | 74.0 (62.8–83.3) | 88/116 | 75.6 (65.8–83.8) | 75.0 (65.5–83.0) | 76.2 (67.7–83.4) | −6.2 (−19.3 to +6.8) | −4.7 (−16.9 to +7.6) |
| Claim: improved care or outcomes | 12/96 | 13.7 (6.3–24.6) | 12/116 | 11.7 (5.4–21.2) | 12.1 (5.8–21.4) | 10.6 (5.1–18.8) | +9.9 (+0.5 to +19.4) | +8.0 (−0.4 to +16.4) |
| Claim beyond predictive performance | 83/96 | 87.7 (80.0–93.1) | 100/116 | 87.3 (80.5–92.4) | 87.1 (80.7–92.0) | 86.8 (81.0–91.4) | +3.7 (−6.4 to +13.8) | +3.3 (−6.5 to +13.1) |
| Claim beyond predictive performance, studies without decision utility or prospective evaluation | 60/71 | 85.9 (76.2–92.7) | 75/89 | 85.4 (77.0–91.7) | 85.4 (77.6–91.4) | 85.1 (78.1–90.6) | – | – |

Primary: strata I, A and B of the PubMed frame. With stratum E: the 352 Embase full reports outside the PubMed frame added as a fourth update stratum (20 eligible studies; weight 352/97 with a finite-population correction). Counts are unweighted; percentages are design-weighted with 95% CIs (Korn–Graubard for prevalences, Wald for the outperformance share). Pooled = legacy census (weight 1) plus the update design. Differences are design-based contrasts, not shown for items met by all or no studies. Studies without a numeric value for both models count as not showing a gain.

**Supplementary Table S14 |** Primary regression of ΔAUROC with and without stratum E.

| Term | Primary, coefficient (95% CI) | Primary, P | With stratum E, coefficient (95% CI) | With stratum E, P |
|---|---|---|---|---|
| Intercept | 0.1056 (0.0486 to 0.1625) | <0.001 | 0.1010 (0.0488 to 0.1533) | <0.001 |
| External validation of the gain (C4) | 0.0086 (−0.0130 to 0.0302) | 0.431 | 0.0152 (−0.0072 to 0.0377) | 0.181 |
| Complete unimodal benchmarking (C2) | −0.0303 (−0.0557 to −0.0049) | 0.020 | −0.0406 (−0.0645 to −0.0167) | 0.001 |
| Log test-set size | −0.0069 (−0.0127 to −0.0012) | 0.019 | −0.0052 (−0.0106 to 0.0002) | 0.060 |
| Period: update (vs legacy) | 0.0126 (−0.0134 to 0.0387) | 0.339 | 0.0108 (−0.0146 to 0.0362) | 0.401 |

Survey-weighted linear regression of ΔAUROC on complete unimodal benchmarking (C2), external validation of the gain (C4), log test-set size and period, in the pooled design. Studies: 110 primary (61 update) and 125 with stratum E (76 update).

**Supplementary Table S15 |** Increase in the reported gain within completely benchmarked studies had a single-source comparator been missing.

| Counterfactual | Group | Studies (n) | Mean increase in ΔAUROC (95% CI) | Median (IQR) | Increase ≥ 0.02, % | Increase ≥ 0.05, % |
|---|---|---|---|---|---|---|
| Strongest source not modelled | Pooled | 78 | 0.062 (0.052 to 0.073) | 0.056 (0.030–0.118) | 83.2 | 47.2 |
| Strongest source not modelled | Legacy | 37 | 0.089 (0.064 to 0.113) | 0.064 (0.031–0.130) | 83.8 | 64.9 |
| Strongest source not modelled | Update | 41 | 0.057 (0.046 to 0.067) | 0.050 (0.030–0.081) | 83.1 | 43.4 |
| One source modelled, chosen at random | Pooled | 78 | 0.035 (0.029 to 0.041) | 0.030 (0.015–0.065) | 59.7 | 28.1 |
| One source modelled, chosen at random | Legacy | 37 | 0.050 (0.036 to 0.064) | 0.034 (0.020–0.075) | 75.7 | 43.2 |
| One source modelled, chosen at random | Update | 41 | 0.032 (0.026 to 0.039) | 0.029 (0.015–0.061) | 56.2 | 24.8 |
| Structured clinical comparator only | Pooled | 60 | 0.044 (0.029 to 0.059) | 0.007 (0.000–0.113) | 48.5 | 30.5 |
| Structured clinical comparator only | Legacy | 31 | 0.061 (0.033 to 0.089) | 0.000 (0.000–0.128) | 45.2 | 38.7 |
| Structured clinical comparator only | Update | 29 | 0.040 (0.022 to 0.057) | 0.027 (0.000–0.081) | 49.4 | 28.4 |

Post hoc. 78 of the 79 analysed studies with complete unimodal benchmarking and AUROC for both models; one study was excluded because the AUROC of one source was reported only in an unavailable supplement. For each constituent source, the best separately fitted single-source model on the primary outcome and test set; 10 studies had three or more sources and 3 used values read from figures. With per-source AUROCs ordered $U(1) \geq U(2) \geq \ldots$, the increase is $U(1) - U(2)$ (strongest source not modelled), $U(1)$ minus the mean of all single-source AUROCs (one source modelled at random) and $U(1)$ minus the AUROC of the structured clinical model, in the 60 studies with one (structured clinical comparator only; the increase is zero in 27 studies where that model was the strongest or tied). Means and shares are design-weighted (pooled design); medians are unweighted. In the pooled sample the mean reported gain was 0.046 and would have been 0.108 without the strongest source's model.